\pdfoutput=1

\documentclass[letterpaper,twocolumn,10pt]{article}
\usepackage{usenix}

\usepackage{tikz}
\usepackage{amsmath}

\usepackage{amsmath,amsfonts,bm}

\def\eqref#1{equation~\ref{#1}}
\def\Eqref#1{Equation~\ref{#1}}

\def\1{\bm{1}}

\DeclareMathAlphabet{\mathsfit}{\encodingdefault}{\sfdefault}{m}{sl}
\SetMathAlphabet{\mathsfit}{bold}{\encodingdefault}{\sfdefault}{bx}{n}

\DeclareMathOperator*{\argmin}{arg\,min}

\usepackage[utf8]{inputenc} 
\usepackage[T1]{fontenc}    
\usepackage{hyperref} 
\usepackage[capitalise]{cleveref}
\usepackage{url}            
\usepackage{booktabs}       
\usepackage{amsfonts}       
\usepackage{nicefrac}       
\usepackage{microtype}      
\usepackage{xcolor}         
\usepackage{wrapfig}
\usepackage{standalone}
\usepackage{bm}
\usepackage{graphicx}
\usepackage{float}
\usepackage{caption}
\usepackage{subcaption}
\usepackage{amsmath}
\usepackage{wrapfig}

\usepackage{multirow}
\usepackage{amssymb}
\usepackage{pifont}
\usepackage{enumitem,kantlipsum}
\usepackage{caption}
\usepackage[usestackEOL]{stackengine}

\usepackage[ruled,vlined]{algorithm2e}
\usepackage{tcolorbox}
\newcommand{\etal}{\textit{et al}. }
\newcommand{\ie}{\textit{i}.\textit{e}., }
\newcommand{\eg}{\textit{e}.\textit{g}. }
\newcommand{\cmark}{\checkmark}
\newcommand{\xmark}{\ding{53}}

\newcommand{\ourmethod}{AutoFHE}
\newcommand{\ourrelu}{EvoReLU}

\newcommand{\fhemp}{MPCNN}
\newcommand{\aespa}{AESPA}
\newcommand{\redsec}{REDsec}

\begin{document}

\date{}

\title{\Large \bf AutoFHE: Automated Adaption of CNNs for Efficient Evaluation over FHE}

\author{{\rm Wei Ao \quad Vishnu Naresh Boddeti}\\
{\normalsize Computer Science and Engineering} \\
{\normalsize Michigan State University} \\
{\tt\small \{aowei, vishnu\}@msu.edu}
}

\maketitle

\begin{abstract}
Secure inference of deep convolutional neural networks (CNNs) under RNS-CKKS involves polynomial approximation of unsupported non-linear activation functions. However, existing approaches have three main limitations: 1) \emph{Inflexibility:} The polynomial approximation and associated homomorphic evaluation architecture are customized manually for each CNN architecture and do not generalize to other networks. 2) \emph{Suboptimal Approximation:} Each activation function is approximated instead of the function represented by the CNN. 3) \emph{Restricted Design:} Either high-degree or low-degree polynomial approximations are used. The former retains high accuracy but slows down inference due to bootstrapping operations, while the latter accelerates ciphertext inference but compromises accuracy. To address these limitations, we present AutoFHE, which automatically adapts standard CNNs for secure inference under RNS-CKKS. The key idea is to adopt layerwise mixed-degree polynomial activation functions, which are optimized jointly with the homomorphic evaluation architecture in terms of the placement of bootstrapping operations. The problem is modeled within a multi-objective optimization framework to maximize accuracy and minimize the number of bootstrapping operations. AutoFHE can be applied flexibly on any CNN architecture, and it provides diverse solutions that span the trade-off between accuracy and latency. Experimental evaluation over RNS-CKKS encrypted CIFAR datasets shows that AutoFHE accelerates secure inference by $1.32\times$ to $1.8\times$ compared to methods employing high-degree polynomials. It also improves accuracy by up to 2.56\% compared to methods using low-degree polynomials. Lastly, AutoFHE accelerates inference and improves accuracy by $103\times$ and 3.46\%, respectively, compared to CNNs under TFHE.
\end{abstract}
\section{Introduction\label{sec:introduction}}

\begin{figure}
    \centering
    \includegraphics[width=\linewidth]{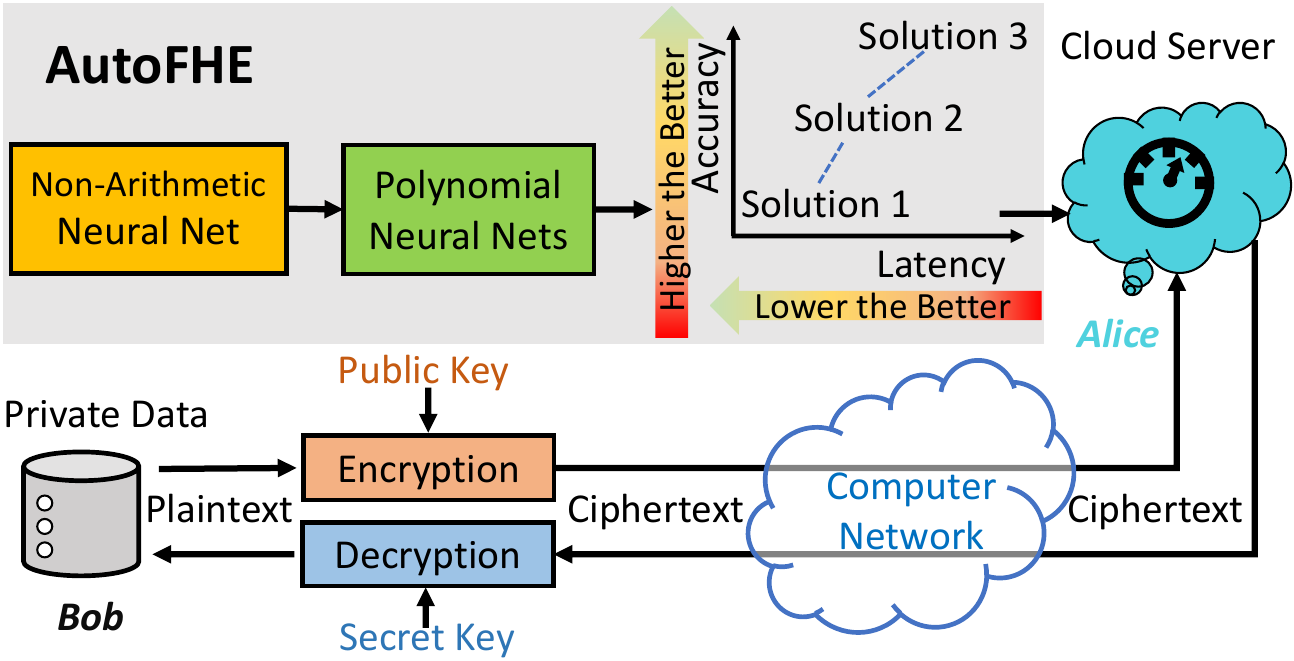}
    \caption{\ourmethod{} can automatically adapt the standard CNN with non-linear activations into a set of polynomial CNNs that span the trade-off between accuracy and latency for ciphertext inference. \ourmethod{} solutions can be deployed on the Cloud server to satisfy a range of customer requirements. \label{fig:overview}}
\end{figure}

\textbf{MLaaS}, machine learning as a service, is a rapidly growing market with many commercial offerings like Amazon Web Services (AWS), Google Google Cloud Platform (GCP), and Microsoft Azure. Its growth has been driven by the widespread success of deep learning on many tasks like vision~\cite{he2016deep,dosovitskiy2020image}, language~\cite{vaswani2017attention,devlin2018bert}, games~\cite{silver2018general,schrittwieser2020mastering}, science~\cite{fawzi2022discovering,jumper2021highly}, and many more. Figure~\ref{fig:overview} shows a typical MLaaS scenario. The Cloud (Alice) holds deep learning models, while the customer (Bob) has private data and requests service from Alice. Bob wants to protect his private data and does not want Alice to learn sensitive information. On the other hand, deep learning models, including \emph{neural architectures} and \emph{trained weights}, are properties of Alice. Alice spends considerable efforts to design neural architectures, like ResNets~\cite{he2016deep}, ViT~\cite{dosovitskiy2020image}, and MLP-Mixer~\cite{tolstikhin2021mlp} and consumes huge computational resources to search for novel neural architectures~\cite{zoph2016neural,liu2018darts} or train network weights~\cite{goyal2017accurate,weng2021large}.

\vspace{3pt}\noindent\textbf{Homomorphic Encryption (HE)}: Secure inference of deep learning models under leveled homomorphic encryption (LHE)\cite{gilad2016cryptonets,brutzkus2019low,lou2021hemet} or fully homomorphic encryption (FHE)~\cite{lee2022privacy,lee2022low,folkerts2023redsec} is a promising approach for resolving security concerns between Bob and Alice in the context of MLaaS. FHE enables us to evaluate a circuit with arbitrary depth, including modern deep CNNs~\cite{lee2022low}. Figure~\ref{fig:overview} shows secure inference of CNNs under FHE. First, Bob generates a public key to encrypt his private data and sends Alice the ciphertext. Second, Alice applies neural networks to process the ciphertext input, yielding an encrypted result. Finally, Bob uses the secret key to decrypt the encrypted result. Under FHE, Bob cannot learn Alice's neural architectures and weights, while Alice is also not exposed to Bob's data or the outcome.

\vspace{3pt}\noindent\textbf{Polynomial CNNs:} Non-arithmetic activation functions, such as $\mathrm{ReLU}(x)=\max(x, 0)$, are a core component of modern CNNs, aiding in learning non-linear decision boundaries between classes. For example, residual networks (ResNets)~\cite{he2016deep} are composed of Conv-BN-ReLU triplets~\cite{he2016deep}. Since FHE only supports \emph{multiplications} and \emph{additions}, ReLU must be replaced by polynomial approximations to evaluate CNNs under FHE. Existing methods to generate polynomial CNNs fall into two categories; \emph{manual design of low-degree and high-degree polynomial approximations}.

\textbf{(1)} A number of approaches adopt low-degree (typically $\leq 3$) polynomials~\cite{gilad2016cryptonets,chou2018faster,brutzkus2019low,mishra2020delphi,lou2020safenet,lou2021hemet,park2022aespa} to substitute non-arithmetic activation functions and \emph{train} the resultant polynomial neural networks \emph{from scratch}. For instance, CryptoNets~\cite{gilad2016cryptonets}, LoLa~\cite{brutzkus2019low} and Delphi~\cite{mishra2020delphi} employ a simple quadratic activation function $x^2$. Faster CryptoNets~\cite{chou2018faster} exploit more accurate low-degree approximation $2^{-3}x^2+2^{-1}x+2^{-2}$. SAFENet~\cite{lou2020safenet} adopts $a_1x^3+a_2x^2+a_3x+a_4$ or $b_1x^2+b_2x+b_3$ and HEMET~\cite{lou2021hemet} uses $ax^2+bx+c$. After low-degree polynomials are plugged into networks like ResNets both \textit{network weights} and \textit{polynomial coefficients} are trained from scratch using stochastic gradient descent (SGD). However, polynomial layers often lead to unstable training since they may dramatically amplify activations during forward propagation and gradients during backward propagation. For example, gradient explosion was observed in prior works~\cite{mishra2020delphi,lou2020safenet}. 
As such, low-degree approaches suffer from a \emph{dilemma}. On the one hand, since low-degree polynomials cannot precisely approximate ReLU, polynomial networks have to be trained from scratch and suffer from poor prediction accuracy. On the other hand, using a higher degree polynomial approximation of ReLU leads to training instability due to exploding gradients. In either case, low-degree approaches achieve lower accuracy than ReLU networks, \eg on CIFAR-10 HEMET~\cite{lou2021hemet} and SAFENet ~\cite{lou2020safenet} report 83.7\% and 88.9\% Top-1 accuracy, respectively. To mitigate gradient explosion, \aespa~\cite{park2022aespa} normalized the outputs of each polynomial basis separately, leading to improved predictive accuracy.

\textbf{(2)} A few approaches use high-degree polynomials to approximate ReLU precisely. So, high-degree approaches do not need to train polynomial networks from scratch and can inherit weights from pretrained ReLU networks. One representative polynomial approximation of ReLU is Minimax composite polynomials~\cite{lee2021minimax,lee2021precise}. By expressing ReLU as $\mathrm{ReLU}(x)=x\cdot(0.5+0.5\cdot \mathrm{sgn}(x))$, a composite polynomial is used to approximate $\mathrm{sgn}(x)$. The approximation of ReLU is defined as $\mathrm{AppReLU}(x)=x\cdot(0.5+0.5\cdot p_{\alpha}(x)),x\in[-1,1]$. $p_{\alpha}(x)$ is the composite Minimax polynomial, and $\alpha$ quantifies approximation precision, \ie $|p_{\alpha}(x)-\mathrm{sgn(x)}|\leq 2^{-\alpha}$.
Given $x\in[-B, B]$, the scaled AppReLU is defined as $B\cdot\mathrm{AppReLU}(x/B)$, with a precision of $B\cdot 2^{-\alpha}$. However, high-degree polynomials consume many multiplicative levels and require numerous bootstrapping operations, leading to a high computational burden. \fhemp~\cite{lee2022low}, the state-of-the-art approach for secure inference of CNNs under RNS-CKKS, adopts Minimax composite polynomials with a precision of $\alpha=13$. By choosing to approximate each ReLU function with high precision, \fhemp{} results in prediction accuracy that is comparable to ReLU networks. However, the same high-degree AppReLU replaces all ReLUs and consumes $\mathbf{\sim50\%}$ levels. The ciphertext quickly exhausts levels and uses bootstrapping to refresh the zero-level ciphertext before every AppReLU layer (refer to Figure~\ref{fig:connect}). As such, the homomorphic evaluation architecture needs to be customized for each CNN architecture. Moreover, bootstrapping operations consume $\mathbf{>70\%}$ of inference time (refer to Table~\ref{tab:runtime}) and result in prohibitively high latency.

\vspace{3pt}\noindent\textbf{Motivation:} We propose \ourmethod{} to address the above-mentioned limitations of existing methods for secure inference of CNNs. It is based on layerwise mixed-degree polynomials and a hybrid of approximation and training methods. The goal is to 
\underline{\textbf{Auto}}matically generate polynomial CNNs and associated homomorphic evaluation architecture spanning a trade-off front of accuracy and latency under \underline{\textbf{FHE}}.

\textbf{(1)} \textsc{Layerwise Mixed-Degree Polynomials:} A plausible solution to decrease the computational burden of secure inference is to assign layerwise mixed-degree polynomials to different ReLUs across a network based on the observation that different layers in a network have varying degree of sensitivity to approximation errors. SAFENet~\cite{lou2020safenet} demonstrated that mixed-degree polynomials can exploit layerwise sensitivity. However, current mixed-degree search frameworks are limited in two aspects. i) \emph{small search space}: \eg SAFENet only provides two polynomials, degree 2 and degree 3. The small search space cannot include all possible solutions from low-degree to high-degree polynomials. ii) \emph{scalarization of multiple objectives}: a weighted sum is used to balance multiple objectives~\cite{mishra2020delphi,lou2020safenet}. However, this approach requires a pre-defined preference to weigh the different objectives. So, they cannot generate diverse solutions to meet different requirements in a single optimization run.

\textbf{(2)} \textsc{Hybrid of Approximation and Training:} Precisely approximating ReLU allows \fhemp~to achieve the state-of-the-art ciphertext inference accuracy. Low-degree approaches train polynomial CNNs from scratch to compensate for loss in accuracy. We posit that by taking advantage of approximation and training, we can inherit weights from ReLU networks and fine-tune polynomial networks to adapt learnable weights to layerwise mixed-degree polynomials. In principle, such a design can allow for high ciphertext inference accuracy while reducing latency.

However, realizing the above goals presents multiple challenges. The design space, which includes layerwise mixed-degree polynomials and the associated homomorphic evaluation architectures in terms of placement of bootstrapping operations, is prohibitively large for effective manual design. Therefore, in this paper, we advocate for automated optimization of the joint polynomial and homomorphic architecture.

\vspace{3pt}\noindent\textbf{Contributions:} We outline our contributions from two perspectives, \emph{design} and \emph{system}.

From a \emph{design} perspective, our contributions are,
\begin{enumerate}[leftmargin=*]
    \item \textsc{Flexibility:} We jointly search for polynomial approximation and a compatible homomorphic evaluation architecture (placement of bootstrapping operations) so we can automatically adapt \emph{any} convolutional neural networks for secure evaluation over FHE.
    \item \textsc{Optimal Approximation:} We approximate the end-to-end function represented by the CNN instead of a standalone non-linear activation function. This allows us to exploit the varying sensitivity of different layers to approximation error, and obtain polynomial approximations with high accuracy and efficient evaluation over FHE.
    \item \textsc{Seamless Design:} We allow for layerwise mixed-degree polynomials, which enables us to find a range of models that span the accuracy and latency trade-off.
\end{enumerate}

From a \emph{system} perspective, our contributions are,
\begin{enumerate}[leftmargin=*]
    \item \textsc{Search Space:} We design search space to include all possible low-degree and high-degree polynomials to enable us to discover better solutions. 
    \item \textsc{Search Objective:} We formulate the search problem as a multi-objective optimization. We automatically generate diverse polynomial networks spanning the trade-off front between accuracy and latency in a single optimization run.
    \item \textsc{Search Algorithms:} We propose combining search and training algorithms to search over large search spaces efficiently, optimize coefficients of arbitrary polynomials, and fine-tune layerwise mixed-degree polynomial networks. Specifically, we propose: 
    \begin{itemize}[leftmargin=*]
    \item \emph{MOS}, a multi-objective search algorithm to search for solutions within a large search space.
    \item \emph{R-CCDE}, a gradient-free search algorithm to optimize coefficients of composite polynomials.
    \item \emph{PAT}, a fine-tuning algorithm that adapts network weights to polynomial activations. 
    \end{itemize}
\end{enumerate}

\noindent\textbf{Experimental Results} on encrypted CIFAR datasets show that \ourmethod~has a better trade-off of accuracy and latency compared to high-degree and low-degree approaches under RNS-CKKS. Compared to high-degree \fhemp~\cite{lee2022low}, \ourmethod~accelerates inference by $\bm{1.32\times\sim1.8\times}$ while improving accuracy by $\bm{+0.08\%\sim 0.3\%}$ on CIFAR10. \ourmethod{} speeds up inference by $\bm{1.1\times\sim1.4\times}$ while increasing accuracy by $\bm{+0.36\%\sim 0.75\%}$ on CIFAR100. Compared to low-degree \aespa~\cite{park2022aespa}, \ourmethod{} improves ciphertext accuracy by $\bm{+2.56\%}$ and $\bm{+2.44\%}$ based on ResNet32 and ResNet44 backbones on CIFAR10 with similar latency. Furthermore, we compare the accuracy-latency trade-off of networks across two FHE schemes, namely RNS-CKKS and TFHE. Specifically, we compare AutoFHE networks designed for RNS-CKKS with \redsec~\cite{folkerts2023redsec}, which is designed for TFHE. We observe that \ourmethod{} improves accuracy by $\bm{+10.06\%}$ and $\bm{+3.46\%}$ compared to \redsec~BNet$_S$ and BNet, respectively, while simultaneously reducing the corresponding latency by $\bm{24\times}$ and $\bm{103\times}$, respectively.

\vspace{3pt}\noindent \textbf{Code}: \url{https://github.com/human-analysis/AutoFHE}.
\section{Preliminaries}
\noindent\textbf{RNS-CKKS: } The full residue number system (RNS) variant of Cheon-Kim-Kim-Song (RNS-CKKS)~\cite{cheon2017homomorphic,cheon2018full} is a leveled homomorphic encryption (LHE) scheme for approximate arithmetic. Under RNS-CKKS, a ciphertext $\bm{c}\in \mathcal{R}_{Q_\ell}^2$ satisfies the decryption circuit $[\langle \bm{c}, sk \rangle]_{Q_\ell}=m+e$, where $\langle \cdot, \cdot \rangle$ is the dot product and $[\cdot]_{Q}$ is the modular reduction function. $\mathcal{R}_{Q_\ell}=\mathbb{Z}_{Q_\ell}[X]/(X^N+1)$ is the residue cyclotomic polynomial ring. The modulus is $Q_\ell=\prod_{i=0}^\ell q_\ell$, where $0 \leq \ell \leq L$. $\ell$ is a non-negative integer, referred to as \emph{level}, which denotes the capacity of homomorphic multiplications. $sk$ is the secret key with Hamming weight $h$. $m$ is the original plaintext message, and $e$ is a small error that provides security. A ciphertext has $N/2$ slots to accommodate $N/2$ complex or real numbers. RNS-CKKS supports homomorphic addition and multiplication:
\begin{align}
    \begin{split}
    \mathrm{Decrypt}( \bm{c} \oplus \bm{c}^\prime) &= \mathrm{Decrypt}(\bm{c})+\mathrm{Decrypt}(\bm{c}^\prime) \approx m + m^\prime \\
    \mathrm{Decrypt} (\bm{c}\otimes \bm{c}^\prime) &= \mathrm{Decrypt}(\bm{c})\times \mathrm{Decrypt}(\bm{c}^\prime) \approx m\times m^\prime
  \end{split}
\end{align}

\vspace{3pt}
\noindent\textbf{Bootstrapping:} LHE only allows a finite number of homomorphic multiplications, with each multiplication consuming one level due to rescaling. Once a ciphertext's level reaches zero, a bootstrapping operation is required to refresh it to a higher level and allow more multiplications. The number of levels needed to evaluate a circuit is known as its \emph{depth}. RNS-CKKS with bootstrapping~\cite{cheon2018bootstrapping} is an FHE scheme that can evaluate circuits of arbitrary depth. It enables us to homomorphically evaluate deep CNNs on encrypted data. Conceptually, bootstrapping homomorphically evaluates the decryption circuit and raises the modulus from $Q_0$ to $Q_L$ by using the isomorphism $\mathcal{R}_{q_0}\cong \mathcal{R}_{q_0} \times \mathcal{R}_{q_1} \times \cdots \times  \mathcal{R}_{q_L}$~\cite{bossuat2021efficient}. Practically, bootstrapping~\cite{cheon2018bootstrapping} homomorphically evaluates modular reduction $[\cdot]_Q$ by first approximating it by a scaled sine function, which is further approximated through polynomials~\cite{cheon2018bootstrapping, lee2020optimal}. Bootstrapping~\cite{bossuat2021efficient} has four stages: ModRaise, CoeffToSlot, EvalMod, and SlotToCoeff. Bootstrapping incurs a lot of key switching operations (KSO), which are the most time-consuming operation on the RNS-CKKS scheme~\cite{lee2022low}. The refreshed ciphertext has level $\ell=L-K$, where $K$ levels are consumed by bootstrapping~\cite{bossuat2021efficient} for polynomial approximation of modular reduction.

\vspace{3pt}\noindent\textbf{Threat Model:} In this paper, we assume the same threat model as prior works under HE, like HEMET~\cite{lou2021hemet}, \fhemp~\cite{lee2022low}, \redsec~\cite{folkerts2023redsec}. As discussed in the MLaaS scenario, a customer uploads encrypted data to a Cloud server and requests ML services. The Cloud uses neural networks to process the ciphertext without decryption and send back an encrypted result. Only the customer holds the secret key and can decrypt the encrypted result. The Cloud cannot learn sensitive information from the customer's data and the result. The customer is also not exposed to the Cloud's neural networks, including architectures and weights.
\section{\ourmethod\label{sec:approach}}
Given a neural network $g(\bm{a}, \bm{\omega}_0(\bm{a}))$ with architecture $\bm{a}$ and \textit{pretrained}\footnote{In our paper, \textit{pretrained} neural networks especially refer to neural networks with ReLU activation.} weights $\bm{\omega}_0(\bm{a})$ and $M$ ReLU layers, \ourmethod~generates polynomial networks on a trade-off front by maximizing accuracy and minimizing latency. For a given architecture $\bm{a}$ during the search, every solution is represented by a triplet of variables $\bm{S}(\bm{a}) = \left(\bm{D}(\bm{a}), \bm{\Lambda}(\bm{a}), \bm{\omega}(\bm{a})\right)$. We will drop the architecture $\bm{a}$ from hereon for ease of notation. $\bm{D}$ is the degree vector of all \ourrelu~layers, $\bm{\Lambda}$ is the coefficients of all \ourrelu~layers, and $\bm{\omega}$ are the trainable weights of the neural network which are initialized with $\bm{\omega}_0$ and fine-tuned for adaptation to the layerwise mix-degree polynomials. We will assign each solution with the minimization objective $\bm{o}(\bm{S})=\{1-\mathrm{Acc}(\bm{S}), \mathrm{Boot}(\bm{S})\}$. $1-\mathrm{Acc}(\bm{S})$ is the validation error, and $\mathrm{Boot}(\bm{S})$ is the number of bootstrapping operations. Note that the accuracy and the number of bootstrapping operations depend on all the variables $\bm{S}$. For instance, appropriately adapting the network weights $\bm{\omega}$ could allow us to use lower-degree polynomial approximations of the activation functions, which reduces the required number of bootstrapping calls.

\ourmethod~is a search-based approach which comprises of three main components: \emph{search space} (\Cref{sec:search_space}), \emph{search objective} (\Cref{sec:search_objective}) and \emph{search algorithm} (\Cref{sec:search_algorithms}). We describe each of these in detail next.

\subsection{Search Space \label{sec:search_space}}

\noindent\textbf{\ourrelu} is a genetic polynomial function used to replace the non-arithmetic ReLU function. We model it as follows,
\begin{equation}
    y = \mathrm{\ourrelu}(x) = \begin{cases}
                x, &\;d=1\\
                \alpha_2x^2 + \alpha_1 x + \alpha_0, &\;d=2\\ 
                x\cdot \left( \mathcal{F}(x) + 0.5 \right), &\; d > 2  \\
                 \end{cases}
    \label{eq:evorelu}
\end{equation}
\noindent where $\mathcal{F}(x)$ is a composite polynomial with $K$ sub-polynomials $f^{d_k}_k$ with degree $d_k$ and seeks to approximate $0.5\cdot \mathrm{sgn}(x)$ for $d>2$.
\begin{equation}
    \mathcal{F}(x)=(f^{d_K}_K\circ \cdots \circ f^{d_k}_{k}\circ \cdots \circ f^{d_1}_1)(x), 1\leq k\leq K
    \label{eq:composite}
\end{equation}

The total degree of $\mathcal{F}(x)$ is $\prod_{k=1}^Kd_k$, and consequently the degree of \ourrelu~is $d=\prod_{k=1}^Kd_k+1$. For $d>2$ the \textbf{multiplicative depth} of \ourrelu~is $ 1 + \sum_{k=1}^{K} \lceil \log_2(d_k + 1) \rceil$ when using the Baby-Step Giant-Step (BSGS) algorithm~\cite{lee2020optimal,bossuat2021efficient} to evaluate composite polynomial $\mathcal{F}(x)$. For $d=1$ and $d=2$, the multiplicative depth is 0 and 2, respectively.

\ourrelu{} is modeled to allow for \emph{automatic} discovery of common solutions in the literature for improving the latency of inference on ciphertexts. For instance, for $d=1$,  $\mathrm{EvoReLU}(x)=x$ is equivalent to removing the corresponding ReLU layer through pruning~\cite{mishra2020delphi,lou2020safenet,jha2021deepreduce}. Similarly, for $d=2$, $\mathrm{EvoReLU}(x)=\alpha_2x^2 + \alpha_1 x + \alpha_0$ is equivalent to using low-degree approximations of ReLU through quadratic functions~\cite{gilad2016cryptonets,brutzkus2019low,chou2018faster,mishra2020delphi,lou2020safenet,lou2021hemet}. Finally, for $d>2$, $\mathrm{EvoReLU}(x)=x\cdot \left( \mathcal{F}(x) + 0.5 \right)$ is equivalent to high-degree polynomial approximations~\cite{lee2021minimax,lee2022low} of ReLU. In this case, \ourrelu~bears similarity to the Minimax composite polynomial in Lee \etal~\cite{lee2021precise, lee2022low}. However, the objective for optimizing the coefficients differs significantly. While Lee~\etal~\cite{lee2021precise, lee2022low} seek to approximate a single ReLU function precisely, our goal is to jointly optimize all EvoReLU functions in a neural network $\bm{a}$ to approximate its corresponding function $g(\bm{a}, \bm{\omega})$.

For the quadratic version of \ourrelu, we allow the coefficients $\alpha_0$, $\alpha_1$, and $\alpha_2$ to differ \emph{channel-wise} when fine-tuning polynomial CNNs. Such a design improves performance and makes optimizing model weights $\bm{\omega}$ more stable. For this case, we also introduce a BatchNorm layer after the quadratic \ourrelu{} for recentering and reshaping the distribution of output activations. We can integrate the BatchNorm parameters into the polynomial coefficients. So, there is no extra consumption of multiplicative levels.

\begin{figure}[!ht]
    \centering
    \includegraphics[width=\linewidth]{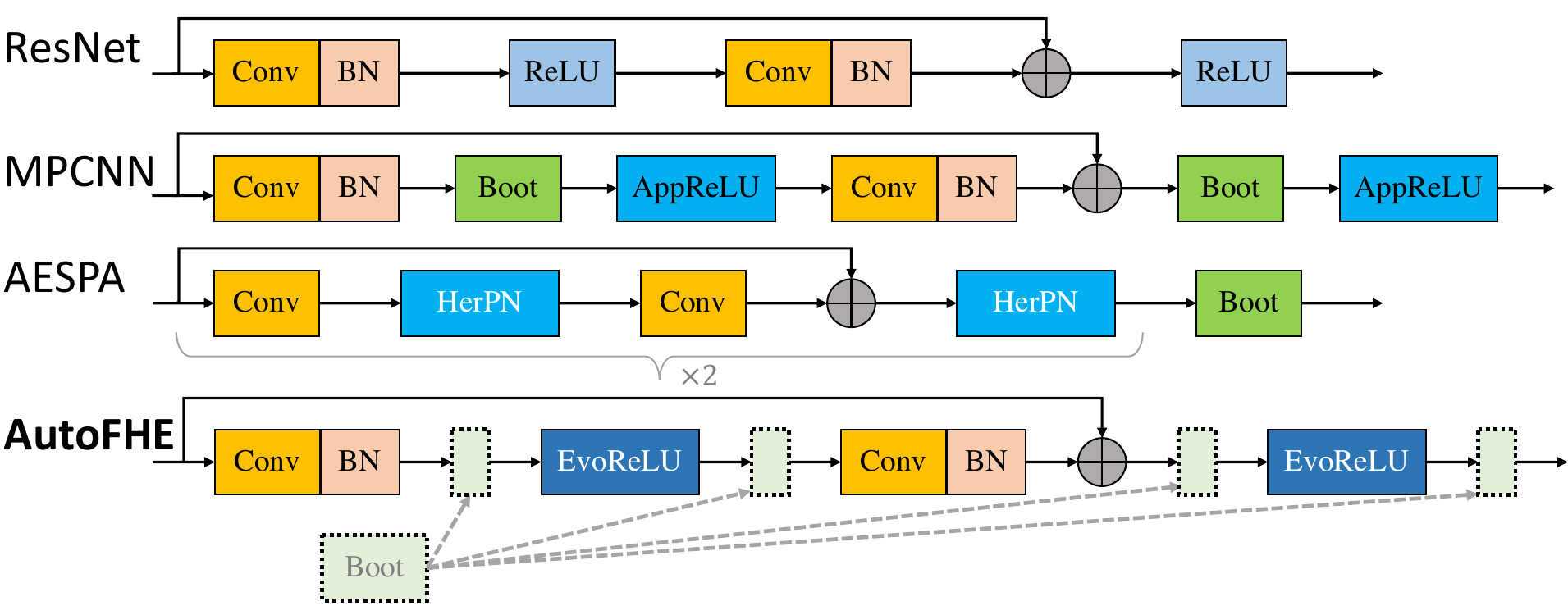}
    \caption{Homomorphic evaluation architectures for Residual Networks (ResNets). 1st row: standard Conv-BN-ReLU triplet~\cite{he2016deep}. 2nd row: \fhemp~\cite{lee2022low} which uses high-degree polynomials. 3rd row: AESPA~\cite{park2022aespa} which uses low-degree polynomials. 4th row: The proposed \ourmethod, which uses layerwise mixed-degree polynomials. Dashed rectangles indicate the plausible locations where bootstrapping can be placed. Both EvoReLU and the placement of bootstrapping operations are searched.\label{fig:connect}}
\end{figure}

We \textbf{represent} the composite polynomial $\mathcal{F}(x)$ by its degree vector $\bm{d}=\{d_k\}_{k=1}^{K}, d_k\in\mathbb{N}$. Each sub-polynomial $f_k^{d_k}(x)$ as a linear combination of Chebyshev polynomials of degree $d_k$,
\begin{equation}
    f^{d_k}_k(x)=\frac{1}{\beta_k} \sum_{i=1}^{d_k} \alpha_i\mathrm{T}_i(x)
    \label{eq:poly}
\end{equation}
\noindent where $\alpha_i\in\mathbb{R}$ and $\beta_k\in\mathbb{R}$. $\mathrm{T}_i(x)$ is the Chebyshev basis of the first kind, $\alpha_i$ are the coefficients for linear combination, and scaling parameter $\beta_k$ is a parameter to scale the output. The coefficients $\bm{\alpha}_k=\{\alpha_i\}_{i=1}^{d_k}$ control the polynomial's shape, while $\beta_k$ controls its amplitude. A composite polynomial with the degree vector $\bm{d}$ has learnable parameters:
\begin{equation}
    \bm{\lambda}=\left\{\bm{\alpha}_1, \beta_1, \cdots, \bm{\alpha}_k, \beta_k, \cdots,  \bm{\alpha}_K, \beta_K \right\} 
\end{equation}

A neural network with $M$ ReLU activations needs $M$ \ourrelu~polynomial activations. $\bm{D}=\{\bm{d}_1, \bm{d}_2, \cdots, \bm{d}_M\}$ is the degree vector of all EvoReLUs and the corresponding coefficient parameters are $\bm{\Lambda} = \{ \bm{\lambda}_1, \bm{\lambda}_2, \cdots, \bm{\lambda}_M \}$.

\vspace{3pt}
\noindent\textbf{Homomorphic Evaluation Architecture} refers to the placement of ConvBN, polynomial, and bootstrapping. In \fhemp, AppReLU depth is 14, ConvBN depth is 2, and the remaining levels after bootstrapping are $L-K=16$. So, bootstrapping is placed after every AppReLU and ConvBN (Figure~\ref{fig:connect}). \aespa~\cite{park2022aespa} uses degree 2 Hermite polynomial (HerPN) with depth 2 to replace ReLU and Batchnorm. Therefore, every 4 Conv-HerPN should be followed by one bootstrapping operation (Figure~\ref{fig:connect}). In \ourmethod, \ourrelu~is layerwise and mixed-degree. The multiplicative depth of \ourrelu{} varies layer by layer. \ourmethod~introduces a flexible evaluation architecture where bootstrapping can be called after ConvBN or \ourrelu~(Figure~\ref{fig:connect}). The flexible evaluation architecture of \ourmethod~can better fit layerwise mix-degree \ourrelu. 
\begin{figure}[!ht]
    \centering
    \includegraphics[width=0.7\linewidth]{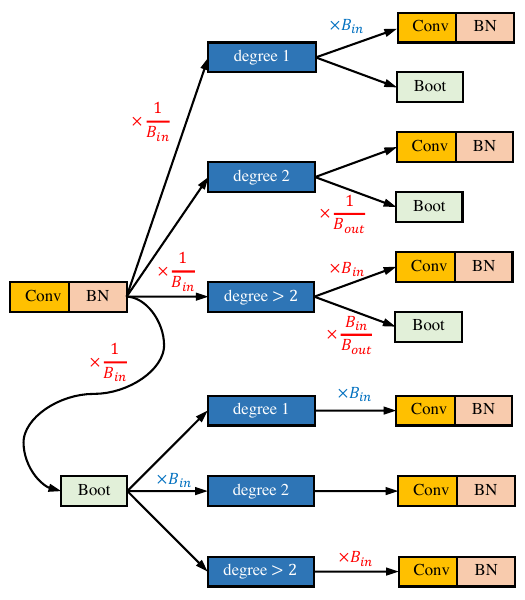}
    \caption{Scaling in \ourmethod. Color Key: \textcolor{red}{integrate scaling into previous operation; \textcolor{blue}{integrate scaling into next operation.}}}
    \label{fig:scaling}
\end{figure}

\vspace{3pt}
\noindent \textbf{Linear Scaling:} Since the domain of Chebyshev polynomials and bootstrapping is $[-1, 1]$, we need to scale the ciphertext to $[-1, 1]$ before Chebyshev polynomials and bootstrapping and reverse it after Chebyshev polynomials and bootstrapping. To avoid consuming levels, scaling is integrated into other operations. In \fhemp, the polynomial neural network with AppReLU can be roughly regarded as \textit{piece-wise linear} due to high precision approximation of high-degree AppReLU. \fhemp~estimates domain of AppReLU using the training dataset and uses the maximum number of all AppReLU domains to scale down input images by $\times1/B$ and scale up ciphertext in the fully connected layer by $\times B$. In \aespa, HerPN uses degree 2 Hermite polynomial and does not need to scale input of HerPN. However, the input of bootstrapping should be scaled to $[-1,1]$. The domain $[-B, B]$ of each bootstrapping can be estimated on the training dataset, then integrate $\times 1/B$ into the previous HerPN and integrate $\times B$ into the next Conv layer. In \ourmethod, bootstrapping can be placed before \ourrelu~and after \ourrelu. So, we need to estimate domain $[-B_{in}, B_{in}]$ and range $[-B_{out}, B_{out}]$ of each \ourrelu~on the training dataset. Figure~\ref{fig:scaling} shows possible scaling scenarios in \ourmethod. We can integrate scaling into the previous or next operation. Specifically, we can multiply convolutional weight, batch norm weight and bias, and polynomial coefficients by the scaling constant. The scaling operation is conducted in plaintext and will not introduce either extra level consumption. With the scaling \ourrelu{} can now be expressed as,
\begin{equation}
    \mathrm{\ourrelu}(x) = \begin{cases}
                x, &\;d=1\\
                \alpha_2x^2 + \alpha_1 x + \alpha_0, &\;d=2\\ 
                x\cdot \left( \mathcal{F}(x/B) + 0.5 \right), &\; d > 2  \\
                 \end{cases}
    \label{eq:evorelu_scaled}
\end{equation}

\begin{table}
    \centering
    \scalebox{0.9}{
    \begin{tabular}{lcc}
    \toprule
    Variable && Option \\
    \midrule
    $\#$ polynomials $(K)$ && 6 \\
    poly degree $(d_k)$ && $\{0,1,3,5,7\}$ \\
    coefficients $(\bm{\Lambda})$ && $\mathbb{R}$ \\
    \bottomrule
    \end{tabular}}
    \caption{Search variables and options. \label{tab:search-options}}    
\end{table}

\begin{table}
    \centering
    \scalebox{0.8}{
    \begin{tabular}{l|ccc}
    \toprule
    Backbone & \#ReLUs & Dimension of $\bm{D}$ & Search Space Size \\
    \midrule \midrule
    ResNet20 & 19 & 114 & $10^{80}$ \\
    ResNet32 & 31 & 186 & $10^{130}$ \\
    ResNet44 & 43 & 258 & $10^{180}$ \\
    VGG11 & 10 & 60 & $10^{42}$ \\
    \bottomrule
    \end{tabular}}
    \caption{Search space of \ourmethod~for ResNet and VGG.\label{tab:seach-space}}
\end{table}

\vspace{3pt}
\noindent \textbf{Search Space:} Our search space includes the number of sub-polynomials ($K)$ in our composite polynomial, the choice of degrees for each sub-polynomial $(d_k)$, and the coefficients of the polynomials $\bm{\Lambda}$. Table \ref{tab:search-options} shows each variable's options. Note that choice $d_k=0$ corresponds to an identity placeholder, so theoretically, the composite polynomial may have fewer than $K$ sub-polynomials. Furthermore, when the degree of $(p_k^{d_k}\circ p_{k-1}^{d_{k-1}})(x)$ is less than or equal to 31 (maximum degree of a polynomial supported on RNS-CKKS~\cite{lee2021minimax,lee2021precise}), we merge the two sub-polynomials into a single sub-polynomial $p_k^{d_k}(p_{k-1}^{d_{k-1}})(x)$ with degree $d_k\cdot d_{k-1}\leq 31$ before computing its depth. This helps reduce the size of the search space and leads to smoother exploration. Table~\ref{tab:seach-space} lists the number of ReLUs of our backbone models and the corresponding dimension and size of search space for $\bm{D}$. Searching for layerwise \ourrelu~is a challenging high-dimensional optimization problem within a vast search space. 

\subsection{Search Objective\label{sec:search_objective}}
\begin{figure*}
    \centering
    \includegraphics[width=\linewidth]{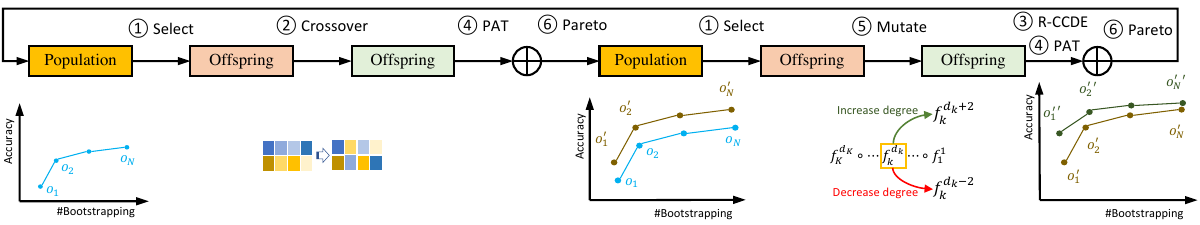}
    \caption{Multi-objective search of \ourmethod. \label{fig:search}}
\end{figure*}

\ourmethod~formulates the search problem as \textit{a multi-objective} optimization
\begin{equation}
    \label{eq:objective}
    \begin{aligned}
    \min_{\bm D} & \quad \left\{1-\mathrm{Acc}_{val}\left(g\left(\bm{\omega}^\ast \mid\bm{D},\bm{\Lambda}(\bm{D})\right)\right),  \mathrm{Boot}(\bm D) \right\} \\
    \mathrm{s.t.} & \quad \bm{\omega}^\ast = \argmin_{\bm{\omega}} \mathcal{L}_{train} \left( g\left(\bm{\omega}\mid\bm{D},\bm{\Lambda}(\bm{D}) \right)\right) \\
    & \quad \mathcal{L}_{train} = (1 - \tau) \mathcal{L}_{CE}+\tau \mathcal{L}_{KL}
    \end{aligned}
\end{equation}

\noindent where $g(\bm{\omega})$ is a neural network with $M$ activation layers and the trainable network weight $\bm{\omega}$. 
The outer multi-objective minimization formulation $\min_{\bm D} \left\{1-\mathrm{Acc}_{val}\left(g\left(\bm{\omega}^\ast\mid\bm{D},\bm{\Lambda}(\bm{D})\right)\right),  \mathrm{Boot}(\bm D) \right\}$ for $\bm{D}$ is to \emph{maximize} the validation accuracy $\mathrm{Acc}_{val}$ as well as \emph{minimize} the number of bootstrapping operations. The coefficient vector $\bm{\Lambda}$ is formulated as a function of $\bm{D}$. In~\Eqref{eq:objective}, $\mathrm{Acc}_{val}$ is the Top-1 accuracy on a validation dataset $val$, $\mathrm{Boot}$ is the number of bootstrapping operations. To determine the number of bootstrapping operations, we count the level consumption of all \ourrelu's to determine where we need to call bootstrapping. By minimizing the number of bootstrapping operations, we search for the placement of bootstrapping and minimize the wasted levels. For example, consider that we have a ciphertext with a level equal to 2, but the next operation consumes 10 levels. We must waste 2 levels and call bootstrapping to refresh the ciphertext first. \ourmethod~can minimize the wasted levels by adjusting the depth of \ourrelu.  $\{\bm{D}_i,\bm{\Lambda}_i\}$ has its corresponding network weight $\bm{\omega}_i$ that can compensate errors introduced by layerwise \ourrelu~$\{\bm{D}_i,\bm{\Lambda}_i\}$. We initialize $\bm{\omega}_i$ with the weight $\bm{\omega}_0$ from the pretrained ReLU network and then fine-tune the network $g(\bm{\omega}_i)$ to minimize the training loss $\mathcal{L}_{train}(\bm{\omega}_i)$ on the training dataset. In summary, the objective in \Eqref{eq:objective} guides the search algorithm to i) explore layerwise EvoReLU, including its \emph{degrees} and \emph{coefficients}; 2) discover the placement of bootstrapping to work well with layerwise mixed-degree \ourrelu; 3) trade-off validation accuracy and inference latency to return diverse polynomial networks. 

The training loss $\mathcal{L}_{train}$ used to optimize the weight $\bm{\omega}$ is $(1 - \tau) \mathcal{L}_{CE}+\tau \mathcal{L}_{KL}$, where $\mathcal{L}_{CE}$ is the cross-entropy loss and  $\mathcal{L}_{KL}$ is the Kullback–Leibler (KL) divergence loss. $\tau$ is a predefined parameter to balance CE and KL loss. In \Eqref{eq:objective}, we omit the variable of $\mathcal{L}_{train}$. Given \ourrelu~$(\bm{D},\bm{\Lambda})$, the variable is weight $\bm{\omega}$. The KL loss computes the distance between distributions of logits of the polynomial network and the ReLU network. We introduce the KL loss because it can push the output of the polynomial network close to the ReLU network. It can be regarded as knowledge distillation (KD)~\cite{hinton2015distilling}. We transfer knowledge from the ReLU network to polynomial networks.

\subsection{Search Algorithms\label{sec:search_algorithms}}
\subsubsection{Multi-Objective Search\label{sec:mos}}
\noindent \textbf{Multi-Objective Optimization:} Given two solutions with minimization objectives $\bm{o}_1,\bm{o}_2\in \mathbb{R}^d$, we want to minimize all items of $\bm{o}_1$ and $\bm{o}_2$. 
If $\bm{o}_{1,i} \leq \bm{o}_{2,i}, \forall i \in \left\{1,2,\cdots,d \right\}$ and 
$\bm{o}_{1,j} < \bm{o}_{2,j}, \exists j \in \left\{1,2,\cdots,d \right\}$, $\bm{o}_1$ \textit{dominates} $\bm{o}_2$~\cite{srinivas1994muiltiobjective,deb2002fast}. It means $\bm{o}_1$ is better than $\bm{o}_2$. It is denoted as $\bm{o}_1 \prec \bm{o}_2$. A set of solutions $\bm{O}=\{\bm{o}_i\}_{i=1}^N$ can be grouped into sub-sets, $\{\bm{o}_i\}_{i=1}^{N_1}$, $\{\bm{o}_i\}_{i=1}^{N_2}$, $\cdots$, corresponding to the 1st trade-off front, the 2nd trade-off front and so on. Solutions within the same trade-off front are not dominated by each other. The 1st trade-off front dominates the 2nd trade-off front, and so on.  

\begin{algorithm}[t]
\caption{MOS\label{algo:mos}}
\SetKwData{Minival}{Minival}\SetKwData{Train}{Train}\SetKwData{Pop}{Population}\SetKwData{Acc}{Acc}\SetKwData{Off}{Offspring}
\SetKwFunction{Lhs}{LHS}\SetKwFunction{Pareto}{Pareto}\SetKwFunction{Crossover}{Crossover}\SetKwFunction{Select}{Select}\SetKwFunction{Mutate}{Mutate}
\SetKwFunction{Eval}{Eval}\SetKwFunction{Rccde}{R-CCDE}\SetKwFunction{Pat}{PAT}
\SetKwInOut{Input}{Input}\SetKwInOut{Output}{Output}\SetKwInOut{Init}{Initial}
\Input{Pretrained Network $g(\bm{\omega}_0)$, population size $N$, offspring size $N^\prime$, number of generations $T$, training dataset \Train, mini-validation dataset \Minival \;}
\Output{Trade-off front $\left\{\bm{D}_i, \bm{\Lambda}_i, \bm{\omega}_i\right\}\Rightarrow\bm{o}_i, 1\leq i\leq N$ \;}
\BlankLine
\For{$t\leftarrow 1$ \KwTo $T$}{
$\{\bm{D}_j,\bm{\Lambda}_j\}_{j=1}^{N^\prime}\gets$ \Select{$\{\bm{D}_i,\bm{\Lambda}_i,\bm{o}_i\}_{i=1}^N$} \;
$\{\bm{D}_j^\prime,\bm{\Lambda}_j^\prime\}_{j=1}^{N^\prime}\gets$ \Crossover{$\{\bm{D}_j,\bm{\Lambda}_j\}_{j=1}^{N^\prime}$} \;
$\{\bm{\omega}_j^\prime\}_{j=1}^{N^\prime}\gets$ \Pat{$\{\bm{\omega}_0; \bm{D}_j^\prime, \bm{\Lambda}_j^\prime\}_{j=1}^{N^\prime}$, \Train} \;
$\{\bm{o}_j^\prime\}_{j=1}^{N^\prime}\gets$ \Eval{$\{\bm{D}_j^\prime, \bm{\Lambda}_j^\prime, \bm{\omega}_j^\prime\}_{j=1}^{N^\prime}$, \Minival} \;
$\left\{\bm{D}_j^\prime, \bm{\Lambda}_j^\prime, \bm{\omega}_j^\prime\right\}\Rightarrow\bm{o}_j^\prime, 1\leq j\leq N^\prime$ \;
$\{\bm{o}_i\}_{i=1}^N\gets$
\Pareto{$\{\bm{o}_i\}_{i=1}^N\cup\{\bm{o}_j^\prime\}_{j=1}^{N^\prime}$} \;
$\left\{\bm{D}_i, \bm{\Lambda}_i, \bm{\omega}_i\right\}\Rightarrow\bm{o}_i, 1\leq i\leq N$ \;
$\{\bm{D}_j,\bm{\Lambda}_j,\bm{\omega}_j\}_{j=1}^{N^\prime}\gets$ \Select{$\{\bm{D}_i,\bm{\Lambda}_i, \bm{\omega}_i,\bm{o}_i\}_{i=1}^N$} \;
$\{\bm{D}_j^\prime\}_{j=1}^{N^\prime}\gets$ \Mutate{$\{\bm{D}_j\}_{j=1}^{N^\prime}$} \;
$\{\bm{\Lambda}_j^\prime\}_{j=1}^{N^\prime}\gets$ \Rccde{$\{\bm{D}_j^\prime\}_{j=1}^{N^\prime}$} \;
$\{\bm{\omega}_j^\prime\}_{j=1}^{N^\prime}\gets$ \Pat{$\{\bm{\omega}_j^\prime;\bm{D}_j^\prime, \bm{\Lambda}_j^\prime\}_{j=1}^{N^\prime}$, \Train} \;
$\{\bm{o}_j^\prime\}_{j=1}^{N^\prime}\gets$ \Eval{$\{\bm{D}_j^\prime, \bm{\Lambda}_j^\prime, \bm{\omega}_j^\prime\}_{j=1}^{N^\prime}$, \Minival} \;
$\left\{\bm{D}_j^\prime, \bm{\Lambda}_j^\prime, \bm{\omega}_j^\prime\right\}\Rightarrow\bm{o}_j^\prime, 1\leq j\leq N^\prime$ \;
$\{\bm{o}_i\}_{i=1}^N\gets$
\Pareto{$\{\bm{o}_i\}_{i=1}^N\cup\{\bm{o}_j^\prime\}_{j=1}^{N^\prime}$} \;
$\left\{\bm{D}_i, \bm{\Lambda}_i, \bm{\omega}_i\right\}\Rightarrow\bm{o}_i, 1\leq i\leq N$ \;
}
\end{algorithm}

\vspace{3pt}\noindent \textbf{MOS:} Figure~\ref{fig:search} and Algorithm~\ref{algo:mos} show our \textbf{M}ulti-\textbf{O}bjective \textbf{S}earch framework for \ourmethod. MOS is an evolutionary search algorithm to solve the multi-objective optimization in~\Eqref{eq:objective}. It maintains a population of solutions distributed on trade-off fronts of accuracy and the number of bootstrapping operations. We crossover and mutate solutions to improve every generation's trade-off fronts, as shown in Figure~\ref{fig:search}. During the search, we define the population size, namely the number of solutions, $N$. These $N$ solutions may be grouped into multiple trade-off fronts, which can improve exploration ability during search. We design operations to search for layerwise mixed-degree polynomials: 

\begin{itemize}
\item[\textcircled{\raisebox{-0.9pt}{1}}] \textsc{Select:} 
Solutions of the current population are first grouped to different trade-off fronts by non-dominated sorting~\cite{deb2002fast}. Solutions on the same trade-off front have the same fitness. The 1st trade-off front has a higher fitness than the 2nd trade-off front, and so on. We apply the tournament selection~\cite{goldberg1991comparative} to improve the diversity of offspring. Specifically, we randomly select three solutions from the current population and keep the solution with the highest fitness. We choose $N^\prime$ solutions from the current population to build the offspring. In our paper, we set $N^\prime=6N$. 

\item[\textcircled{\raisebox{-0.9pt}{2}}] \textsc{Crossover} enables network-level information exchange. As shown in Figure~\ref{fig:search}, we randomly and uniformly exchange \ourrelu~layers between two solutions (parents) to generate two new solutions. However, new solutions cannot inherit weights from parent solutions because weights are adapted to the parent's layerwise mixed-degree polynomials. So, we initialize weights with pretrained weights and then fine-tune new solutions. 

\item[\textcircled{\raisebox{-0.9pt}{3}}] \textsc{R-CCDE} optimizes \ourrelu~coefficients.

\item[\textcircled{\raisebox{-0.9pt}{4}}] \textsc{PAT} fine-tunes new polynomial CNNs.

\item[\textcircled{\raisebox{-0.9pt}{5}}] \textsc{Mutation} is to locally explore \ourrelu. We randomly increase or decrease the degree to smoothly change polynomials, as shown in Figure~\ref{fig:search}. We randomly increase or decrease the degree of a sub-polynomial of \ourrelu{} with predefined probabilities. 

\item[\textcircled{\raisebox{-0.9pt}{6}}] \textsc{Pareto} refers to non-dominated sorting and crowding distance sorting~\cite{deb2002fast}. We apply Pareto to select $N$ solutions from both population and offspring ($N+N^\prime$ solutions) to build a new population ($N$ solutions). 
\end{itemize}

\subsubsection{R-CCDE \label{sec:rccde}}
\begin{algorithm*}[ht]
\SetKwFunction{Lhs}{LHS}
\SetKwInOut{Input}{Input}\SetKwInOut{Output}{Output}\SetKwInOut{Init}{Initial}
\caption{R-CCDE\label{algo:rccde}}
\Input{Composite polynomial $\mathcal{F}(x)=(f_K^{d_K} \circ f_{k-1}^{d_{k-1}} \circ \cdots \circ f_1^{d_1})(x)$ with parameters $\bm{\lambda}=\{\bm{\alpha}_1, \beta_1, \cdots, \bm{\alpha}_k , \beta_k, \cdots \bm{\alpha}_K, \beta_K\}$, target function $q(x)$, number of generations $T$, scaling decay $\gamma$ \;}
\Output{Context vector $\bm{\lambda}^\ast=(\bm{\alpha}_1^\ast, \beta_1^\ast, \cdots ,\bm{\alpha}_k^\ast, \beta_k^\ast, \cdots, \bm{\alpha}_K^\ast, \beta_K^\ast)$\;}
\Init{$\bm{\lambda}^\ast\gets$ \Lhs{$\sum_{k=1}^Kd_k+K$}}
\BlankLine
\For{$t\leftarrow 1$ \KwTo $T$}{
\For{$k\leftarrow 1$ \KwTo $K$}
{${\color{red}{\bm{\alpha}_k^\star}}\gets \argmin_{\color{blue}\bm{\alpha}_k}\mathcal{L}_{\mathcal{F},q}({\color{blue}\bm{\alpha}_k}|\bm{\lambda}^\ast)\quad\mathrm{s.t.}\quad{\color{blue}\bm{\alpha}_k}|\bm{\lambda}^\ast=(\bm{\alpha}_1^\ast, \beta_1^\ast, \cdots, {\color{blue}\bm{\alpha}_k}, \cdots, \bm{\alpha}_K^\ast, \beta_K^\ast)$\;
$\bm{\lambda}^\ast \gets (\bm{\alpha}_1^\ast, \beta_1^\ast, \cdots, {\color{red}\bm{\alpha}_k^\star}, \cdots, \bm{\alpha}_K^\ast, \beta_K^\ast)$\;
${\color{red}\beta_k^\star} \gets \argmin_{\color{blue}\beta_k}\mathcal{L}_{\mathcal{F},q}({\color{blue}\beta_k}|\bm{\lambda}^\ast)+\gamma\cdot {\color{blue}\beta_k}^2\quad\mathrm{s.t.}\quad{\color{blue} \beta_k}|\bm{\lambda}^\ast=(\bm{\alpha}_1^\ast, \beta_1^\ast, \cdots, {\color{blue}\beta_k}, \cdots, \bm{\alpha}_K^\ast, \beta_K^\ast)$\;
$\bm{\lambda}^\ast \gets (\bm{\alpha}_1^\ast, \beta_1^\ast, \cdots, {\color{red}\beta_k^\star}, \cdots, \bm{\alpha}_K^\ast, \beta_K^\ast)$\;
}}
\end{algorithm*}

\noindent \textbf{Coevolution:} The composite polynomial used by \ourrelu~is: $y_1=f_1^{d_1}(x|\bm{\alpha}_1,\beta_1)$, $\cdots$, $y_K=f_K^{d_K}(y_{K-1}|\bm{\alpha}_K, \beta_K)$. The forward architecture of the composite polynomial, $x\mapsto y_1 \mapsto y_2 \cdots \mapsto y_{K-1} \mapsto y$ is suitable for \emph{coevolution}~\cite{yang2008large,mei2016competitive,ma2018survey} and provides a natural \textit{decomposition}. We can sequentially adjust every sub-polynomial to push the output $y$ close to the target non-arithmetic function. Given the degree $\bm{d}$, the learnable parameter of \ourrelu~ $\bm{\lambda}=(\bm{\alpha}_1, \beta_1, \cdots, \bm{\alpha}_K, \beta_K)$ is grouped into  $\{\bm{\alpha}_1\}, \{\beta_1\}, \cdots, \{\bm{\alpha}_K\}, \{\beta_K\}$. The coefficient $\bm{\alpha}$ controls the shape of the sub-polynomial output, while the scaling parameter $\beta$ controls the amplitude. We sequentially update $\{\bm{\alpha}_k\}$ followed by $\beta_k$, $1\leq k\leq K$, since i) sub-polynomials close to input will have a larger effect on the output, and ii) it is easier to learn coefficients by decoupling the amplitude from the coefficients.

\vspace{3pt}\noindent \textbf{Differentiable Evolution:} The \ourrelu~variables $\{\bm{\alpha}_k\}_{k=1}^K$ and $\{\beta_k\}_{k=1}^K$ are in the continuous space. We adopt a simple yet effective search algorithm to optimize these variables. Differentiable evolution (DE)~\cite{rauf2021adaptive} only uses the \emph{difference} between solutions to optimize continuous variables. Given the following minimization problem in the continuous space
\begin{equation}
    \bm{x}^\ast = \argmin_{\bm{x}} \mathcal{F}\left(\bm{x}\right)
\end{equation}
where $\bm{x}\in\mathbb{R}^d$ and $\mathcal{F}$ is the minimization objective. DE maintains a set of solutions $\bm{X}=\{\bm{x}_i\}_{i=1}^N, \bm{x}_i\in\mathbb{R}^d$. The mutation, crossover, and selection of DE are defined as: 
\begin{equation}
    \begin{aligned}
        \mbox{Mutation:}\; & \bm{v} = \bm{x}_i + F \cdot \left( \bm{x}_j - \bm{x}_k \right), 1 \leq i,j,k \leq N \\
        \mbox{Crossover:}\; & \bm{u}[t] = \begin{cases}
                \bm{v}[t],&\mathcal{U}(0, 1) \leq CR\\
                 \bm{x}_i[t],& \text{Otherwise}
        \end{cases}, 1\leq t \leq d \\
        \mbox{Selection:}\; & \bm{u} = \begin{cases}
                \bm{u},&\mathcal{F}(\bm{u}) \leq \mathcal{F}(\bm{x}_i)\\
                 \bm{x}_i,& \text{Otherwise}
        \end{cases}
    \end{aligned}
    \label{eq:de}
\end{equation}
\noindent where $F\in\mathbb{R}$ is the scaling factor, $CR\in\mathbb{R}$ is the crossover rate, and $\mathcal{U}(0,1)$ is the uniform distribution between 0 and 1. \Eqref{eq:de}~shows a simple strategy to update solutions by only using difference. First, mutation updates $\bm{x}_i$ with the scaled difference $F \cdot \left( \bm{x}_j - \bm{x}_k \right)$. Then, we randomly select items from $\bm{v}$ or $\bm{x}_i$ to generate a new solution $\bm{u}$. Finally, we evaluate $\mathcal{F}(\bm{u})$ and use $\bm{u}$ to replace $\bm{x}_i$ if $\mathcal{F}(\bm{u}) \leq \mathcal{F}(\bm{x}_i)$. DE only uses difference and does not suffer from gradient exploding. It maintains a set of solutions and is not sensitive to initialization. 

\vspace{3pt}\noindent\textbf{R-CCDE:} We propose \textbf{R}egularized \textbf{C}ooperative \textbf{C}oevolution \textbf{D}ifferentiable \textbf{E}volution, called R-CCDE, to search for parameters of $\mathrm{EvoReLU}(x,\bm{\lambda};\bm{d})$, namely $\bm{\lambda}=\left\{\bm{\alpha}_k,\beta_k\right\}_{k=1}^K$. The scaling parameters $\left\{\beta_k\right\}_{k=1}^K$ are used to adjust the amplitude of sub-polynomials during the search. After the search, $\left\{\beta_k\right\}_{k=1}^K$ will be used to scale $\left\{\bm{\alpha}_k\right\}_{k=1}^K$ and obtain coefficients of polynomials. The decomposition makes the search easier by decoupling the shape and amplitude of polynomials. We detail the implementation of R-CCDE in Algorithm~\ref{algo:rccde}. R-CCDE takes as input a composite polynomial $\mathcal{F}(x)=(f_K^{d_K} \circ f_{k-1}^{d_{k-1}} \circ \cdots \circ f_1^{d_1})(x)$ with parameters $\bm{\lambda}=\{\bm{\alpha}_1, \beta_1, \cdots \bm{\alpha}_k , \beta_k \cdots \bm{\alpha}_K, \beta_K\}$. Because EvoReLU is defined as $y=\mathrm{EvoReLU}(x)=x\cdot\left(\mathcal{F}(x) + 0.5\right)$ in \Eqref{eq:composite}, we use the composite polynomial $\mathcal{F}(x)$ to approximate $q(x)=0.5\cdot\mathrm{sgn}(x)$. We set the number of generations and the scaling decay parameter to $T$ and $\gamma$, respectively. The objective function $\mathcal{L}_{\mathcal{F},q}(\cdot)$ is the $\ell_1$ distance between the composite polynomial $\mathcal{F}(x)$ and the target function $q(x)$. R-CCDE maintains a \emph{context vector}~\cite{mei2016competitive} $\bm{\lambda}^\ast = (\bm{\alpha}_1^\ast, \beta_1^\ast, \cdots, \bm{\alpha}_K^\ast, \beta_K^\ast)$ as the best solution so far. $\bm{\lambda}^\ast$ is initialized via Latin hypercube sampling (LHS). In Algorithm~\ref{algo:rccde}, $\bm{\alpha}_k$ and $\beta_k$ $1\leq k\leq K$ are optimized using DE sequentially and alternatively. In generation $t$, given the $k$-th position, we optimize $\bm{\alpha}_k$ as
\begin{equation}
    \begin{aligned}
        {\color{red}{\bm{\alpha}_k^\star}}&= \argmin_{\color{blue}\bm{\alpha}_k}\mathcal{L}_{\mathcal{F},q}({\color{blue}\bm{\alpha}_k}|\bm{\lambda}^\ast)\\
        \mathrm{s.t.}\quad{\color{blue}\bm{\alpha}_k}|\bm{\lambda}^\ast&=(\bm{\alpha}_1^\ast, \beta_1^\ast, \cdots {\color{blue}\bm{\alpha}_k}, \cdots, \bm{\alpha}_K^\ast, \beta_K^\ast)
    \end{aligned}
    \label{eq:rccde1}
\end{equation}
\noindent where $\color{blue}\bm{\alpha}_k$ is a variable, while other $\bm{\alpha}$'s and $\beta$'s are fixed. A candidate solution of $\color{blue}\bm{\alpha}_k$ is plugged into $\bm{\lambda}^\ast$. Then, we evaluate the candidate solution $\color{blue}\bm{\alpha}_k$ by evaluating $\mathcal{L}_{\mathcal{F},q}\left( \bm{\alpha}_1^\ast, \beta_1^\ast,\cdots,{\color{blue} \bm{\alpha}_k},\cdots,\bm{\alpha}_K^\ast, \beta_K^\ast \right)$. We adopt DE to solve the single-objective optimization problem in the continuous space. We maintain a set of candidate solutions of $\bm{\alpha}_k$, namely $\bm{X}=\left\{\bm{x}_i\right\}_{i=1}^N,\bm{x}_i\in\mathbb{R}^{d_k}$. Mutation, crossover, and selection defined in \Eqref{eq:de} are applied to update solutions in $\bm{X}$. Then, the best solution in $\bm{X}$ is assigned to $\color{red}\bm{\alpha}_k^\star$. We use $\color{red}\bm{\alpha}_k^\star$ to replace $\bm{\alpha}_k^\ast$ in the context vector $\bm{\lambda}^\ast$ to update $\bm{\lambda}^\ast$
\begin{equation}
    \bm{\lambda}^\ast = (\bm{\alpha}_1^\ast, \beta_1^\ast, \cdots {\color{red}\bm{\alpha}_k^\star}, \cdots, \bm{\alpha}_K^\ast, \beta_K^\ast)
\end{equation}
\noindent In summary, i)$\{\bm{\alpha}_k\}_{k=1}^K$ and $\{\beta_k\}_{k=1}^K$ \textit{separately} maintain their sets of solutions that are optimized by DE; ii) the \textit{context} vector $\bm{\lambda}^\ast$ is not only the best solution so far. It allows different variables to share information. When evolving $\{\beta_k\}_{k=1}^K$, the objective introduces a \textit{regularization} term
\begin{equation}
    \begin{aligned}
        {\color{red}\beta_k^\star} = \argmin_{\color{blue}\beta_k}\underbrace{\mathcal{L}_{\mathcal{F},q}({\color{blue}\beta_k}|\bm{\lambda}^\ast)}_{\ell_1~\text{Distance}}\quad+\underbrace{\gamma\cdot {\color{blue}\beta_k}^2}_\text{Regularization}\\
        \mathrm{s.t.}\quad{\color{blue}\beta_k}|\bm{\lambda}^\ast=(\bm{\alpha}_1^\ast, \beta_1^\ast, \cdots, {\color{blue}\beta_k}, \cdots, \bm{\alpha}_K^\ast, \beta_K^\ast)
    \end{aligned}
    \label{eq:rccde2}
\end{equation}
\noindent where $\gamma\cdot\beta_k^2$ is the regularization term and $\gamma$ is the scaling decay parameter. Without the regularization $\gamma\cdot\beta_k^2$, we observe $\{\beta_k\}_{k=1}^K$ prefers \textit{large} numbers. Because $p^{d_k}_k(x)=\frac{1}{\beta_k} \sum_{i=1}^{d_k} \alpha_i\mathrm{T}_i(x)$, large $\{\beta_k\}_{k=1}^K$ numbers can make the composite polynomial \textit{numerical stable} because the polynomial output is scaled to a small number. However, it is hard to distinguish different solutions of $\{\bm{\alpha}_k\}_{k=1}^K$. By introducing the regularization term $\gamma\cdot\beta_k^2$, DE prefers large numbers in earlier generations and gradually reduces $\beta_k$. Therefore, DE is not biased toward solutions with large $\beta_k$ numbers. We use R-CCDE to optimize coefficients of quadratic and high-degree \ourrelu. For quadratic \ourrelu~in \Eqref{eq:evorelu}, $\alpha_2$ is obtained by R-CCDE and $\alpha_1=0.5$, $\alpha_0=0$. 

\subsubsection{Polynomial-Aware Training \label{sec:pat}}
Replacing ReLU with \ourrelu~in pretrained neural networks injects \textit{minor} approximation errors, which leads to performance loss. Fine-tuning can mitigate this performance loss by allowing the learnable weights (e.g., convolution or fully connected layers) to adapt to the approximation error. However, backpropagation through \ourrelu~easily leads to exploding gradients since the gradients may be amplified exponentially due to many composite polynomials. From \Eqref{eq:evorelu}, high-degree \ourrelu~can precisely approximate ReLU, while ReLU pruning (degree 1) and quadratic \ourrelu~(degree 2) have a larger approximation error. For low-degree \ourrelu~(degree $\leq2$), we can use SGD to compute gradients because they do not amplify gradients. For high-degree \ourrelu~(degree $>2$), we can use gradients from the original non-arithmetic ReLU function for \emph{backpropagation}. Specifically, during \emph{the forward} pass, \ourrelu~injects slight errors captured by the training loss. During the \emph{backward} pass, we bypass high-degree \ourrelu~and use ReLU to compute gradients to update the weights of the linear trainable layers. We refer to this procedure as \textbf{P}olynomial-\textbf{A}ware \textbf{T}raining (PAT). 
PAT is inspired by STE~\cite{bengio2013estimating} and QAT~\cite{jacob2018quantization}, which uses two different functions for forward- and back-propagation. PAT is defined as:
\begin{equation}
    \frac{\partial \mathrm{\ourrelu}(x)}{\partial x} = \begin{cases}
                1, &\;d=1\\
                2\alpha_2x + \alpha_1, &\;d=2\\ 
                \partial \mathrm{ReLU}(x) / \partial x, &\; d > 2  \\
                 \end{cases}
    \label{eq:grad}
\end{equation}
\section{Experiments\label{sec:experiments}}

\textbf{Datasets:} We benchmark \ourmethod~on CIFAR10 and CIFAR100~\cite{krizhevsky2009learning}. Both datasets have 50,000 training and 10,000 validation images at a resolution of $32\times32$. CIFAR10 has 10 classes, while CIFAR100 includes 100 classes. 
The validation images are treated as private data and used only for evaluating the final networks. 
To guide the search process, we randomly select 10,000 images from the training split as a \emph{minival}~\cite{tan2021efficientnetv2} dataset and use the Top-1 accuracy on the minival dataset to optimize~\Eqref{eq:objective}. In addition, PAT uses the training split to fine-tune polynomial networks. Finally, as our final result, we report the Top-1 accuracy on the encrypted validation dataset under RNS-CKKS.

\vspace{3pt}
\noindent\textbf{Parameters:} 
\textbf{(1)} \textsc{Training Parameters:} We train ReLU networks (used by \fhemp~and \ourmethod) and \aespa~using SGD optimizer with batch size 128, epochs 200, learning rate 0.1, momentum 0.9 and weight decay 0.0005. We use a cosine learning rate scheduler. We clip gradients to 1 when we train polynomial networks (\aespa{} or \ourmethod{}). 
\textbf{(2)} \textsc{Search Parameters:} For MOS, we set the number of generations to 10. Population size is proportional to the number of variables. We set population size to 10, 20, 30 and 40 for VGG11, ResNet20, ResNet32 and ResNet44, respectively. The offspring size is $6\times$ as the population size. When we mutate a polynomial, its degree is decreased by $-2$ with a probability of 0.5 and is increased by $+2$ with a probability of 0.3. For R-CCDE, we set the search domain of $\bm{\alpha}$ to $[-5, 5]$ and that of $\beta$ to $[1, 5]$. We use the set of 20 solutions for optimizing $\beta$. For $\bm{\alpha}$, we set the number of solutions equal to $20\times$ the number of variables. We set both the scaling factor $F$ and the crossover rate $CR$ to 0.5.
We set the scaling decay to $\gamma=0.01$ and the number of iterations to 100. We run R-CCDE 10 times with different random seeds and retain the best solution. 
\textbf{(3)} \textsc{Finetuning Parameters:} 
To fine-tune \ourmethod~polynomial networks, we train them using PAT with batch size 128, learning rate 0.02, momentum 0.9, weight decay 0.0005, and KL weight $\tau=0.9$. We clip gradients to 1. During the search, to quickly estimate the accuracy of polynomial networks, we set epochs to 5. After the search, we set epochs to 90 and use the cosine annealing learning rate scheduler.
\textbf{(4)} \textsc{Cryptographic Parameters:} We followed \fhemp~to set the same cryptographic parameters~\cite{lee2022low} of RNS-CKKS for \fhemp, \aespa~and \ourmethod. The cyclotomic polynomial uses degree $N=2^{16}$. The Hamming weight of the secret key is 192. The ciphertext level is $L=30$, while bootstrapping uses 14 levels ($K=14$). Base modulus, special modulus, and bootstrapping modulus are set to 51 bits, while default modulus is set to 46 bits~\cite{lee2022low}. The cryptographic parameters satisfy \emph{128-bit security}~\cite{lee2022low,cheon2019hybrid}. 

\vspace{3pt}\noindent\textbf{Hardware and RNS-CKKS Library:} \textbf{(1)} \textsc{Search:} On one NVIDIA RTX A6000 GPU, the search process for ResNet-20/32/44 and VGG11 on CIFAR10 took 44 hours, 64 hours, 88 hours, 13 hours, respectively. The search for ResNet32 and VGG11 on CIFAR100 took 67 and 12 hours, respectively. To accelerate R-CCDE, we use 100 CPU threads of AMD EPYC 7502 32-core processor. 
\textbf{(2)} \textsc{FHE Inference:} We evaluate latency under FHE on the publicly available Amazon AWS instance, r5.24xlarge, which has 96 CPU threads and 768 GB RAM. We build C++ implementation of \ourmethod~under RNS-CKKS on top of \fhemp~using Microsoft SEAL library~\cite{seal3.6}. We adopt \fhemp~implementations of Conv, BN, Downsample, AvgPool, and FC layers. 

\begin{table}
    \centering
    \scalebox{0.66}{
    \begin{tabular}{ccccccc}
         \toprule
         Method & Venue & Scheme & Polynomial & \small{Layerwise} & \small{Strategy} & Arch \\
         \midrule \midrule
         \fhemp\cite{lee2022low} & ICML22 & CKKS & high-degree & \xmark & approx & manual  \\
         \aespa\cite{park2022aespa} & arXiv22 & CKKS & low-degree & \xmark & train & manual \\ 
         \redsec\cite{folkerts2023redsec} & NDSS23 & TFHE & n/a  & n/a & train & manual \\
         \textbf{\ourmethod} & USENIX24 & CKKS & mixed & \cmark & adapt & search \\
         \bottomrule
    \end{tabular}}
    \caption{\ourmethod~and baselines. \ourmethod, \fhemp~and \aespa~use fixed-point arithmetic under RNS-CKKS, while \redsec~adpots ternary neural networks under TFHE.}
    \label{tab:baseline}
\end{table}
\begin{figure}[!ht]
    \centering
    \includegraphics[width=0.9\linewidth]{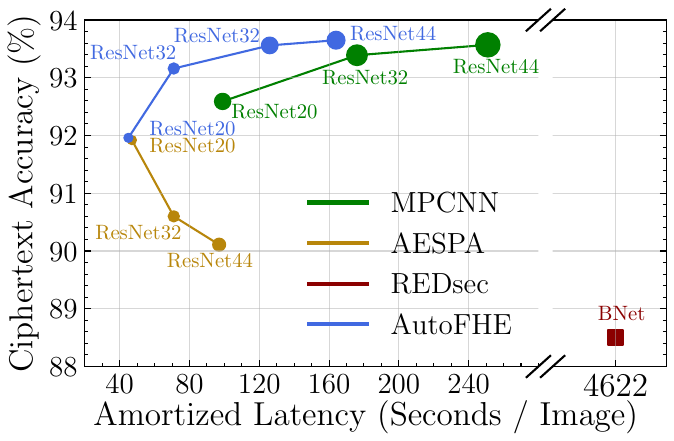}
    \caption{Comparison of trade-offs between ciphertext accuracy and amortized latency on encrypted CIFAR10. The latency is evaluated on Amazon AWS r5.24large using 96 threads. RNS-CKKS approaches take 96 encrypted images as input, while \redsec~processes one image at a time. Circle size is proportional to the number of bootstrapping operations. \label{fig:tradeoff_all}}
\end{figure}

\vspace{3pt}\noindent\textbf{Baselines:} We compare the proposed \ourmethod~with two recent state-of-the-art approaches under RNS-CKKS, high-degree polynomial and approximation-based approach \fhemp~\cite{lee2022low} and low-degree polynomial and training-based approach \aespa~\cite{park2022aespa}, as shown in Table~\ref{tab:baseline}. We also benchmark against \redsec~\cite{folkerts2023redsec} under TFHE to compare different FHE schemes. 

\textsc{RNS-CKKS Baselines:} \fhemp~uses high-degree Minimax composite polynomial approximation of ReLU and reports high ciphertext accuracy (refer to Appendix~\ref{appendix:mpcnn}). \aespa~applies degree 2 Hermite polynomial to replace ReLU, which reduces the multiplicative depth of polynomials and greatly reduces the consumption of bootstrapping. Because \aespa~was originally proposed under secure MPC, we implement \aespa~under RNS-CKKS (refer to Appendix~\ref{appendix:aespa}). We use the same training parameters for \fhemp, \aespa~and \ourmethod. We estimate scaling parameters of \fhemp~on training datasets: 21.26 (ResNet20), 21.99 (ResNet32), 17.80 (ResNet44), 29.82 (VGG11) on CIFAR10 dataset and 63.40 (ResNet32) and 54.97 (VGG11) on CIFAR100. 

\textsc{TFHE Baseline:} The fast fully homomorphic encryption scheme over the torus (TFHE)~\cite{chillotti2020tfhe} provides very fast bootstrapping by using bootstrapped binary gates. \redsec{} applies efficient ternary networks (TNNs) under TFHE. \redsec~reports performance of BNet$_S$ and BNet on CIFAR10 under CPU TFHE~\cite{tfhe1.1}. BNet$_S$ and BNet observe ciphertext accuracy $81.9\%$ and $88.5\%$, and latency 1,081 and 4,622 seconds per image on CIFAR10 dataset on AWS r5.24large instance using 96 CPU threads~\cite{folkerts2023redsec}. The TFHE cryptographic parameters used by \redsec{} satisfy 128 bits of security~\cite{folkerts2023redsec}. Please note that \redsec~uses a different parallel acceleration strategy. \redsec~takes advantage of all 96 CPU threads to process one encrypted image, while RNS-CKKS approaches allocate one thread to each image. Therefore, \redsec{} has the same latency and amortized latency. When we evaluate the latency of \ourmethod, \fhemp~and \aespa~on AWS r5.24large, we input 96 encrypted images using 96 CPU threads. The amortized latency is $\frac{1}{96}\times$ as the latency under RNS-CKKS. So, the amortized latency of \redsec~and RNS-CKKS approaches is comparable.

\subsection{\ourmethod~under RNS-CKKS}
\begin{figure*}[!ht]
    \centering
    \includegraphics[width=0.8\linewidth]{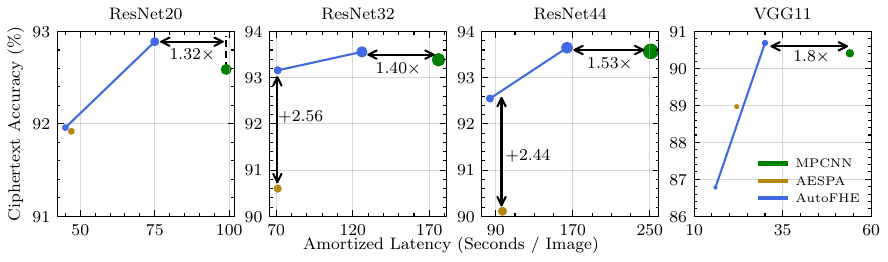}
    \caption{Trade-offs between ciphertext accuracy and amortized latency of ResNet and VGG backbones on CIFAR10.}
    \label{fig:tradeoffs}
\end{figure*}

\begin{table*}[!ht]
\begin{center}
\scalebox{0.725}{
\begin{tabular}{l|ccc|cccc|cccc|cccc}
    \toprule
    \multirow{3}{*}{Dataset} & \multicolumn{3}{c|}{Backbone} & \multicolumn{4}{c|}{MPCNN~\cite{lee2022low}} & \multicolumn{4}{c|}{AESPA\cite{park2022aespa}} & \multicolumn{4}{c}{\textbf{\ourmethod}} \\
    & Network & \small{Params} &  \small{\Centerstack{Plain \\ Acc\%}} & \small{\Centerstack{Bootst-\\rapping}} & \small{\Centerstack{Cipher \\ Acc\% }} & Latency & \small{\Centerstack{Amortized \\ Latency}} & \small{\Centerstack{Bootst-\\rapping}}  & \small{\Centerstack{Cipher \\ Acc\%  }} & Latency & \small{\Centerstack{Amortized \\ Latency}} & \small{\Centerstack{Bootst-\\rapping}}  & \small{\Centerstack{ Cipher\\Acc\%}} & Latency & \small{\Centerstack{Amortized \\ Latency}} \\
    \midrule \midrule
    \multirow{8}{*}{CIFAR10} & \multirow{2}{*}{ResNet20} & \multirow{2}{*}{269K} & \multirow{2}{*}{92.66} & \multirow{2}{*}{18}  & \multirow{2}{*}{92.59}  & \multirow{2}{*}{9,481s}  & \multirow{2}{*}{99s}  & \multirow{2}{*}{\textcolor{brown}{5}}  & \multirow{2}{*}{91.92}  & \multirow{2}{*}{4,554s}  & \multirow{2}{*}{47s} & \textcolor{brown}{5} & 91.96 & \textcolor{blue}{4,295s} & \textcolor{blue}{45s}  \\
    &  &  &  & &  &  &  &  &  &  &  & 11 &  \textcolor{red}{92.89} & 7,191s & 75s \\
    \cmidrule{2-16}
    & \multirow{2}{*}{ResNet32} & \multirow{2}{*}{464K} & \multirow{2}{*}{93.46} & \multirow{2}{*}{30}  & \multirow{2}{*}{93.39}  & \multirow{2}{*}{16,910s}  & \multirow{2}{*}{176s}  & \multirow{2}{*}{\textcolor{brown}{8}}  & \multirow{2}{*}{90.60}  & \multirow{2}{*}{\textcolor{blue}{6,841s}}  & \multirow{2}{*}{\textcolor{blue}{71s}} & \textcolor{brown}{8} & 93.16 & 6,860s & \textcolor{blue}{71s} \\
    &  &  &  & &  &  &  &  &  &  &  & 19 & \textcolor{red}{93.56} & 12,111s & 126s \\
    \cmidrule{2-16}
    & \multirow{2}{*}{ResNet44} & \multirow{2}{*}{658K} & \multirow{2}{*}{93.72} & \multirow{2}{*}{42}  & \multirow{2}{*}{93.57}  & \multirow{2}{*}{24,082s}  & \multirow{2}{*}{251s}  & \multirow{2}{*}{11}  & \multirow{2}{*}{90.11}  & \multirow{2}{*}{9,345s}  & \multirow{2}{*}{97s} & \textcolor{brown}{8} &  92.55 & \textcolor{blue}{8,078s} & \textcolor{blue}{84s} \\
    &  &  &  & &  &  &  &  &  &  &  & 22 & \textcolor{red}{93.65} & 15,769s & 164s \\
    \cmidrule{2-16}
    & \multirow{2}{*}{VGG11} & \multirow{2}{*}{123K} & \multirow{2}{*}{90.53} & \multirow{2}{*}{9}  & \multirow{2}{*}{90.41}  & \multirow{2}{*}{5,221s}  & \multirow{2}{*}{54s}  & \multirow{2}{*}{2}  & \multirow{2}{*}{88.97}  & \multirow{2}{*}{2,112s}  & \multirow{2}{*}{22s} &\textcolor{brown}{1} & 86.78 & \textcolor{blue}{1,515s} & \textcolor{blue}{16s} \\
    &  &  &  & &  &  &  &  &  &  &  & 4 & \textcolor{red}{90.69} & 2,879s & 30s \\
    \midrule
    \multirow{2}{*}{CIFAR100} &  \multirow{1}{*}{ResNet32} & \multirow{1}{*}{472K} & \multirow{1}{*}{70.81}  & \multirow{1}{*}{30} & \multirow{1}{*}{70.59}  &  \multirow{1}{*}{15,693s}  &  \multirow{1}{*}{163s} & \multirow{1}{*}{\textcolor{brown}{8}} & \multirow{1}{*}{68.43}  &  \multirow{1}{*}{\textcolor{blue}{7,171s}}  &  \multirow{1}{*}{\textcolor{blue}{75s}} & 16 & \textcolor{red}{71.34} & 10,969s & 114s \\
     \cmidrule{2-16}
     &  \multirow{1}{*}{VGG11} & \multirow{1}{*}{129K} & \multirow{1}{*}{64.64}  & \multirow{1}{*}{9} & \multirow{1}{*}{63.95}  &  \multirow{1}{*}{5,148s}  &  \multirow{1}{*}{54s} & \multirow{1}{*}{\textcolor{brown}{2}} & \multirow{1}{*}{62.90}  &  \multirow{1}{*}{\textcolor{blue}{2,225s}}  &  \multirow{1}{*}{\textcolor{blue}{23s}} & 7 & \textcolor{red}{64.31} & 4,562s  & 48s \\
    \bottomrule
\end{tabular}
}
\caption{\ourmethod~under the RNS-CKKS scheme. Latency for 96 images is evaluated on AWS r5.24large using 96 threads. Amortized latency is the average latency of each image. We report the ciphertext accuracy of 10,000 encrypted validation images. For CIFAR10, we select two solutions with different bootstrapping for each backbone network. Color key: \textcolor{red}{highest ciphertext accuracy}; \textcolor{blue}{lowest (amortized) latency}; \color{brown}{smallest number of bootstrapping operations}. \label{tab:result}}
\end{center}
\end{table*}

\begin{table*}[!ht]
\begin{center}
\scalebox{0.685}{
\begin{tabular}{l|c|cccc|cccc|cccc}
    \toprule
    \multirow{2}{*}{Dataset} & \multirow{2}{*}{Backbone} & \multicolumn{4}{c|}{MPCNN~\cite{lee2022low}} & \multicolumn{4}{c|}{AESPA\cite{park2022aespa}} & \multicolumn{4}{c}{\textbf{\ourmethod}} \\
    & & \small{\#Boot} & Linear(s) & \small{AppReLU(s)} & Boot(s)  & \small{\#Boot} & Linear(s) & \small{HerPN(s)} & Boot(s)  & \small{\#Boot} & Linear(s) & \small{\ourrelu(s)} & Boot(s)\\
    \midrule \midrule
    \multirow{8}{*}{CIFAR10} & \multirow{2}{*}{ResNet20} & \multirow{2}{*}{18}  & \multirow{2}{*}{$1180\pm32$}  & \multirow{2}{*}{$1067\pm45$}  & \multirow{2}{*}{ $7138\pm95$ }  & \multirow{2}{*}{5}  & \multirow{2}{*}{ $2344\pm9$}  & \multirow{2}{*}{$54\pm1$}  & \multirow{2}{*}{$2108\pm14$} & 5 &  $2014\pm11$ &  $115\pm3$  &   $2092\pm37$ \\
     &  & &  &  &  &  &  &  &  & 11 & $2142\pm52$ & $388\pm18$  &  $4473\pm71$ \\
    \cmidrule{2-14}
    & \multirow{2}{*}{ResNet32} & \multirow{2}{*}{30}  & \multirow{2}{*}{$2104\pm84$}  & \multirow{2}{*}{ $1708\pm103$}  & \multirow{2}{*}{ $12304\pm506$}  & \multirow{2}{*}{8}  & \multirow{2}{*}{ $3582\pm173$}  & \multirow{2}{*}{ $79\pm2$}  & \multirow{2}{*}{$3234\pm76$} & 8 & $3540\pm41$  &  $38\pm1$ &  $3107\pm57$ \\
    &  & &  &  &  &  &  &  &  & 19 & $2962\pm95$ &  $646\pm20$  & $8185\pm142$\\
    \cmidrule{2-14}
    & \multirow{2}{*}{ResNet44} & \multirow{2}{*}{42}  & \multirow{2}{*}{ $3084\pm93$}  & \multirow{2}{*}{ $2385\pm59$}  & \multirow{2}{*}{ $18071\pm433$}  & \multirow{2}{*}{11}  & \multirow{2}{*}{$4683\pm19$}  & \multirow{2}{*}{$110\pm2$}  & \multirow{2}{*}{$4393\pm105$} & 8 &  $4595\pm21$  &  $42\pm1$ &  $3265\pm94$  \\
    &  & &  &  &  &  &  &  &  & 22 & $4447\pm152$ &  $806\pm41$  &  $10256\pm184$ \\
    \cmidrule{2-14}
    & \multirow{2}{*}{VGG11}  & \multirow{2}{*}{9} & \multirow{2}{*}{$676\pm29$}  & \multirow{2}{*}{ $639\pm24$}  & \multirow{2}{*}{$3809\pm92$}  & \multirow{2}{*}{2}  & \multirow{2}{*}{$1281\pm50$}  & \multirow{2}{*}{ $28\pm1$} &  \multirow{2}{*}{$835\pm8$} & 1 & $1056\pm6$  &  $14\pm1$  &  $402\pm21$ \\
    &  & &  &  &  &  &  &  &  & 4 &  $1019\pm7$ &  $172\pm3$ &  $1661\pm4$ \\
    \midrule
    \multirow{2}{*}{CIFAR100} &  ResNet32 & 30 &  $2062\pm64$ &  $1621\pm51$ & $11815\pm123$ & 8  &   $3754\pm20$ &  $79\pm2$  & $3225\pm78$  & 16 &   $3602\pm25$ &  $620\pm16$ &  $6528\pm131$ \\
     \cmidrule{2-14}
     &  VGG11 & 9 &  $685\pm17$  & $645\pm24$ &  $3752\pm82$  & 2 &  $1330\pm6$  & $28\pm1$ &  $835\pm9$ & 7 &  $1145\pm15$ &  $338\pm18$  &  $3027\pm39$  \\
    \bottomrule
\end{tabular}
}
\caption{Latency of operations under the RNS-CKKS scheme. We report the mean and standard deviation of latency of 96 images in Table~\ref{tab:result}. Linear layers include Conv, BN, Downsample, AvgPool, and FC. \label{tab:runtime}}
\end{center}
\end{table*}

\textbf{Trade-offs under FHE:} Figure~\ref{fig:tradeoff_all} shows trade-offs between ciphertext accuracy and amortized latency on CIFAR10 dataset. We benchmark the performance of ResNet and VGGNet backbones on CIFAR10 and CIFAR100 datasets, as shown in Table~\ref{tab:result}. For TFHE baseline \redsec, it reported $81.9\%$ ciphertext accuracy with latency 1,081s per image and $88.5\%$ ciphertext accuracy with latency 4,622s per image on CIFAR10~\cite{folkerts2023redsec}. We have the following observations: 

\begin{tcolorbox}[title={RNS-CKKS \textit{vs} TFHE},halign=left,valign=center]
    Neural networks under RNS-CKKS significantly outperform those under TFHE in terms of both ciphertext accuracy and latency.
\end{tcolorbox}

From Figure~\ref{fig:tradeoff_all}, \redsec~shows lower ciphertext accuracy because ternary neural networks (TNNs) enormously compress models~\cite{li2016ternary} compared to real-valued models used by RNS-CKKS approaches. Real-valued networks have much better representation learning ability than TNNs. In secure inference of neural networks, \redsec~has to evaluate millions to billions of gates~\cite{folkerts2023redsec}. Although evaluating one gate is very fast ($10\sim13$ ms)~\cite{folkerts2023redsec} on TFHE~\cite{tfhe1.1}, the latency of the whole network is still extremely high. The most efficient solution (ResNet20 with 5 bootstrapping operations) of \textbf{\ourmethod}~reports $\bm{91.96\%}$ ciphertext accuracy and \textbf{45s} latency. Compared to \redsec~BNet$_S$ and BNet, \ourmethod{} improves ciphertext accuracy by $\bm{+10.06\%}$ and $\bm{+3.46\%}$, respectively, while being $\bm{24\times}$ and $\bm{103\times}$ faster in terms of amortized latency.

\begin{tcolorbox}[title={High-Degree \textit{vs} Low-Degree \textit{vs} Mixed-Degree},halign=left,valign=center]
    High-degree polynomials are suitable for shallow and deep neural networks but suffer from high latency. Low-degree polynomials can effectively accelerate ciphertext inference but at the cost of a significant drop in accuracy. Mixed-degree layerwise polynomials (\ourmethod) lead to a significantly better trade-off between accuracy and latency.
\end{tcolorbox}

From Figure~\ref{fig:tradeoff_all} and Table~\ref{tab:result}, we observe that high-degree \fhemp~is able to preserve plaintext accuracy although it consumes more levels (14) for AppReLU. Since \fhemp~focuses on approximating the ReLU function, it~can use weights from ReLU networks and does not suffer from exploding activations and gradients in the forward and backward passes of data through the network. In summary, \fhemp~is a plug-in approach with high accuracy and high latency. Low-degree \aespa~reduces level consumption of each polynomial by $12$ compared to \fhemp. \aespa~polynomials only consume $26\%$ and $22\%$ of the bootstrapping operations in \fhemp~for ResNet and VGG, respectively. However, \aespa~achieves this at the cost of large drops in accuracy, $0.7\%\sim3.6\%$ on CIFAR10 and $1.7\%\sim2.4\%$ on CIFAR100 compared to the corresponding plaintext backbone models.

\ourmethod~takes advantage of layerwise mixed-degree polynomials to reduce bootstrapping consumption while maintaining high accuracy. From Figure~\ref{fig:tradeoffs} and Table~\ref{tab:result}, \ourmethod~shows a better trade-off between ciphertext accuracy and latency than \fhemp~and \aespa. Compared to \fhemp~on CIFAR10, \textbf{\ourmethod}~accelerates encrypted image inference by $\bm{1.32\times\sim 1.8\times}$ while improving accuracy by $\bm{+0.08\%\sim 0.3\%}$ in comparison to \fhemp. Similarly, on CIFAR100, \textbf{\ourmethod}~speeds up inference by $\bm{1.1\times\sim 1.4\times}$ while increasing accuracy by $\bm{+0.36\%\sim 0.75\%}$. From Figure~\ref{fig:tradeoffs}, compared to \aespa, \ourmethod{} improves ciphertext accuracy by $\bm{+2.56\%}$ and $\bm{+2.44\%}$ for ResNet32 and ResNet44 backbones on CIFAR10 with similar amortized latency. However, \aespa{}'s low-degree polynomials lead to an increasing drop in accuracy with model depth. This starkly contrasts with plaintext models, where deeper models are known to improve performance. \ourmethod{} can improve performance with depth while maintaining the same latency as \aespa{}. As such, we observe from Figure~\ref{fig:tradeoff_all} and Table~\ref{tab:result} that \ourmethod{} enjoys a much better trade-off between accuracy and latency.

In summary, the challenge of navigating the vast joint design space of polynomial approximations of non-linear activation functions and homomorphic evaluation architectures limits manual approaches like \fhemp{} and \aespa{} to simplify solutions like approximating non-linear activation functions and uniform placement of bootstrapping operations. In contrast, \ourmethod{} algorithmically navigates the design space and identifies solutions that significantly \emph{pareto dominate} manual approaches in accuracy, latency, or both.

\begin{tcolorbox}[title={Approximation \textit{vs} Training \textit{vs} Adaptation},halign=left,valign=center]
    High-precision function approximation can preserve plaintext accuracy without training, while low-precision function approximation with training leads to a loss in accuracy. \ourmethod{} is a hybrid method that inherits the representation learning ability of ReLU networks and adapts the network's learnable weights to layerwise polynomials.
\end{tcolorbox}

On the one hand, \fhemp~has a drop in accuracy of $0.07\sim0.15\%$ and $0.22\sim0.69\%$ compared to plaintext backbone models on CIFAR10 and CIFAR100, respectively. This demonstrates that high-degree polynomials still introduce slight approximation errors. On the other hand, \aespa{} has a significant drop in accuracy, especially for deeper networks, which was also observed by \aespa's authors \cite{park2022aespa}. The results demonstrate that the representation learning ability of (low-degree) polynomial neural networks is inferior to ReLU networks~\cite{leshno1993multilayer}. Unlike \aespa, \ourmethod~inherits the representation learning ability from ReLU networks by using pretrained weights and transferring knowledge. Furthermore, we fine-tune polynomial networks using very small learning rates to adapt learnable network weights to layerwise mixed-degree polynomial \ourrelu. Therefore, \ourmethod{} can achieve the \emph{high-accuracy} of ReLU networks and the \emph{low-latency} of low-degree polynomial networks. As such, compared to \fhemp~and \aespa, \ourmethod~improves both prediction accuracy and reduces inference latency over all ResNet and VGG backbones.

\vspace{3pt}
\noindent\textbf{Operations under RNS-CKKS:} Table~\ref{tab:runtime} shows the latency of different operations under RNS-CKKS. AppReLU is a high-degree polynomial, so its evaluation latency is higher than degree 2 HerPN. The latency of \ourrelu{} is roughly between AppReLU and HerPN. Low-bootstrapping solutions of \ourmethod~further speed up evaluation of polynomial compared to HerPN, \emph{e.g.} ResNet32 with eight bootstrapping, ResNet44 with eight bootstrapping and VGG11 with one bootstrapping. In \fhemp, bootstrapping dominates latency with $74\sim77\%$ of total inference time. In \aespa, linear layers and bootstrapping operations consume similar runtime.

The latency of \ourmethod{} is similar to \aespa~for low-bootstrapping solutions and to \fhemp~for high-bootstrapping solutions. Linear operations of \fhemp~are faster than \aespa~and \ourmethod{} since they are being evaluated at a lower level. For example, \fhemp~ConvBN always takes level 2 ciphertext as input. The evaluation of low-level ciphertexts is faster than high-level ciphertexts. For \aespa~and \ourmethod, polynomial activations (HerPN and \ourrelu) have smaller multiplicative depth, and their linear operations take ciphertexts at higher levels as input. From Table~\ref{tab:runtime}, we observe that i) \ourmethod~can effectively accelerate neural network inference on RNS-CKKS by removing time-consuming bootstrapping operations; ii) Layerwise mixed-degree \ourrelu~effectively explores how to reduce the multiplicative depth of polynomials and further decrease consumption of bootstrapping operations. 

\begin{figure*}[!ht]
    \centering
    \includegraphics[width=0.88\linewidth]{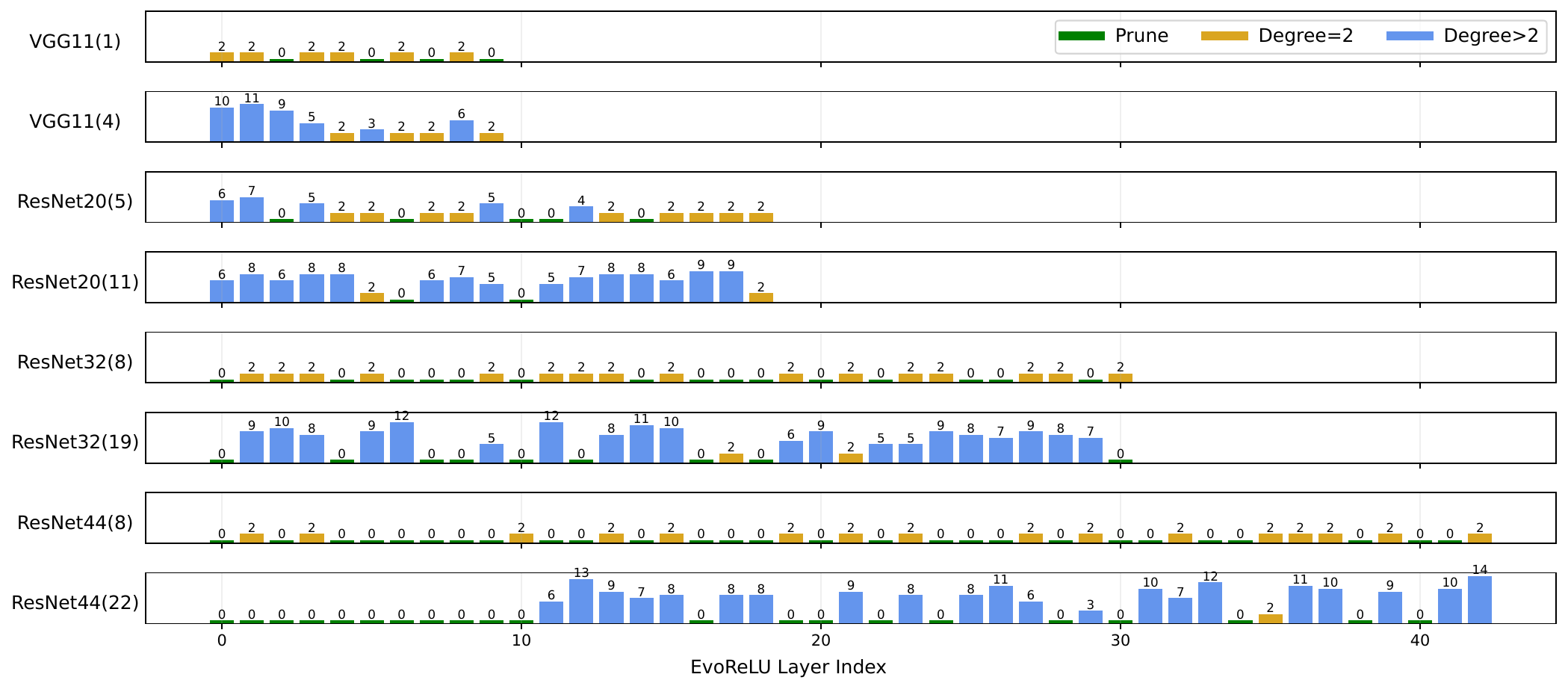}
    \vspace{-0.25cm}
    \caption{Multiplicative depth of layerwise mixed-degree \ourrelu~layers. \label{fig:depth}}
    
\end{figure*}
\begin{figure}[!ht]
    \centering
    \includegraphics[width=\linewidth]{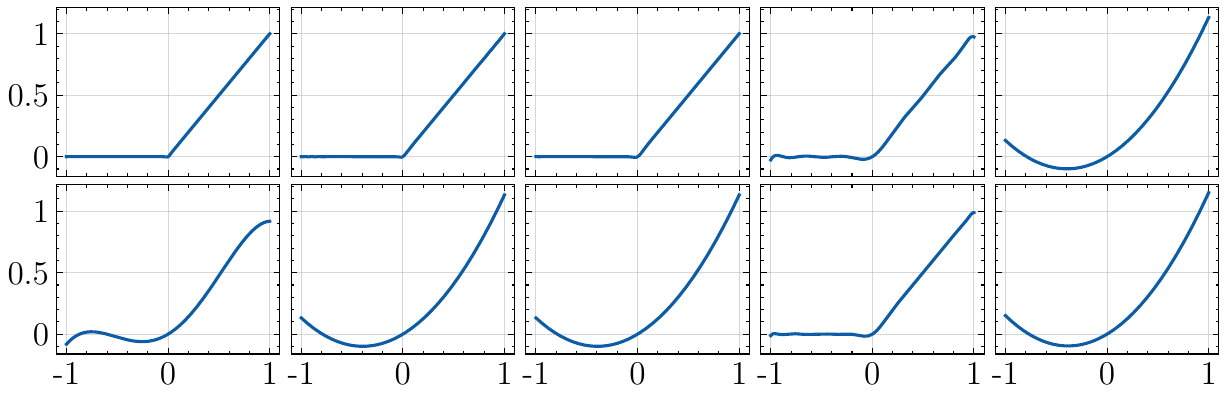}
    \caption{\ourrelu~functions of \ourmethod{} for VGG11 with 4 bootstrapping operations. The top row shows \ourrelu{} for layer $0\sim4$, and the bottom row shows layer $5\sim9$. \label{fig:evorelu_vgg11}}
\end{figure}

\subsection{Layerwise \ourmethod}
\textbf{Depth Distribution:} To analyze the layerwise mixed-degree \ourrelu~discovered by \ourmethod, we study (see Figure~\ref{fig:depth}) the distributions of multiplicative depth for different backbones on CIFAR10. In contrast to the uniform allocation used by \fhemp~and \aespa, the optimal layerwise allocation of \ourrelu{} discovered by \ourmethod{} is highly non-uniform. Such a distribution is challenging to design manually, thus further motivating the need for automated design of layerwise mixed-degree polynomial approximations of activation functions. From the distribution in Figure~\ref{fig:depth}, we identify the following design principles that can guide the design of polynomial neural networks under RNS-CKKS. 

\begin{tcolorbox}[title={Observation 1},halign=left,valign=center]
    Low and high bootstrapping solutions share a similar distribution of multiplicative depth. 
\end{tcolorbox}

In Figure~\ref{fig:depth}, we provide two solutions with low and high bootstrapping operations for each backbone. These two solutions share similar depth distributions. Specifically, high-degree polynomials are preferred in the same layers of low and high bootstrapping solutions. 

\begin{tcolorbox}[title={Observation 2},halign=left,valign=center]
    Depth distribution is linearly scalable. 
\end{tcolorbox}

Consider the depth distributions of VGG11(4) and ResNet20(5), ResNet20(11) and ResNet32(19) in Figure~\ref{fig:depth}. They have different numbers of layers. If we scale their distributions horizontally to match the number of layers, their depth distributions are very similar. This demonstrates that the number of layers and position of activations are the most important factors affecting the sensitivity of layerwise approximation. Therefore, the depth distributions are linearly scalable to the neural network's depth.

\begin{tcolorbox}[title={Observation 3},halign=left,valign=center]
    Consecutive linear layers can be integrated into a single operation to reduce multiplicative depth.
\end{tcolorbox}

Many networks have consecutive linear (depth 0) layers, especially ResNet44(22). In this case, it is possible to integrate successive linear layers into a single linear layer, which decreases the multiplicative depth and removes bootstrapping operations. Table~\ref{tab:runtime} demonstrates that reducing linear operations is an effective way to further accelerate inference.

\vspace{3pt}
\noindent\textbf{Layerwise \ourrelu:} Figure~\ref{fig:evorelu_vgg11} shows layerwise mixed-degree \ourrelu~functions for VGG11 with 4 bootstrapping operations. High-degree \ourrelu~functions can precisely approximate ReLU, \emph{e.g.} layer 0, 1 and 2. Medium-degree \ourrelu~functions (layer 3, 5, 8) observe oscillation but still are very close to ReLU. Degree 2 \ourrelu~is a quadratic function and introduces more approximation errors compared to high-degree and medium-degree functions. For high-degree and medium-degree polynomials, input will be scaled to $[-1, 1]$ so we can prevent exploding activations. Figure~\ref{fig:evorelu_vgg11} qualitatively shows that high-degree and medium-degree \ourrelu~functions can precisely approximate ReLU, and we can use gradients of ReLU in PAT to prevent exploding gradients during backpropagation. Since degree 2 \ourrelu~has a relatively big approximation error, we use SGD rather than approximate gradients from ReLU (refer to \Eqref{eq:grad}).
\section{Related Work \label{sec:related}}

In this paper, we focus on secure inference under FHE. Secure multiparty computation (MPC) is an alternative approach for secure inference~\cite{liu2017oblivious,juvekar2018gazelle,mishra2020delphi,lou2020safenet,ghodsi2021circa,knott2021crypten,rathee2021sirnn}. It is usually employed in a hybrid protocol involving both MPC and HE. MPC primitives include secret sharing, Yao's garbled circuits ~\cite{yao1986generate,bellare2012foundations}, and Beaver's multiplicative triples~\cite{beaver1995precomputing}, \textit{etc.} For example: (1) Gazelle~\cite{juvekar2018gazelle} adopts packed additively homomorphic encryption (PAHE) to evaluate linear layers (Conv and FC layers) and garbled circuits to evaluate non-linear layers (ReLU and MaxPooling layers). (2) Delphi~\cite{mishra2020delphi} uses Garbled circuits to evaluate ReLU, and Beaver's multiplicative triples to evaluate the polynomial approximation $x^2$ of ReLU. (3) Iron~\cite{hao2022iron} employs the Brakerski-Fan-Vercauteren (BFV) scheme~\cite{brakerski2012fully,fan2012somewhat} for matrix multiplication. Non-linear operations, like SoftMax, GELU, and LayerNorm, are evaluated through secret sharing. 

Secure MPC-based approaches for secure inference must carefully consider the trade-off between computation and communication~\cite{juvekar2018gazelle} since both the customer and the Cloud perform computations. In some practical scenarios, sufficient communication and computing resources on the client side may not be available. Pure FHE-based approaches provide customers a \emph{fire-and-forget}~\cite{folkerts2023redsec} service, where they are not involved in the computations and simply wait for the encrypted result. However, FHE-based approaches may have higher latency and memory footprint than secure MPC approaches.

In terms of polynomial neural networks, FHE approaches have to replace \emph{all} ReLU activations with polynomials since FHE only supports multiplications and additions. Secure MPC approaches usually replace only a fraction of ReLUs to reduce online communication and computation costs and retain a few ReLU layers to preserve accuracy, \eg Delphi~\cite{mishra2020delphi} and SAFENet~\cite{lou2020safenet}. These approaches, however, only use low-degree polynomials and observe a significant drop in accuracy when most ReLU activations are replaced. \ourmethod{} can also be employed for secure MPC schemes by changing the search objective from the number of bootstrapping operations to online communication or computation costs of secure MPC.
\section{Conclusion\label{sec:conclusion}}
Non-interactive end-to-end inference of homomorphically encrypted ciphertext images over convolutional networks is an attractive solution for mitigating the security and privacy concerns of cloud-based MLaaS offerings. Adapting CNNs for inference over FHE ciphertexts presents several challenges, including the optimal design of polynomial approximations of non-linear activation functions and associated homomorphic evaluation architecture. Existing solutions primarily rely on manual designs, which are neither scalable nor flexible enough to be applied to any architecture and cater to the needs of different MLaaS customers.

To overcome these challenges, this paper introduced \ourmethod{}, an automated approach for adapting any convolutional neural network for secure inference under RNS-CKKS. It is a multi-objective search algorithm that generates a set of polynomial networks and their associated homomorphic evaluation architecture under FHE by trading off accuracy and latency. It exploits layerwise mixed-degree polynomial activations across different layers in a network and jointly searches for placement of bootstrapping operations for evaluation under RNS-CKKS. We designed a custom search space for layerwise mixed-degree polynomials and adopted multiple objectives for optimization. We also proposed a combination of search and training algorithms, including multi-objective search algorithm \emph{MOS}, composite polynomial coefficient optimization method \emph{R-CCDE}, and polynomial aware network training strategy \emph{PAT}. We extensively evaluate \ourmethod{} on ResNets and VGGNets over encrypted CIFAR datasets. Compared to high-degree \fhemp, \ourmethod{} accelerates inference by $1.32\times\sim 1.8\times$. Compared to low-degree \aespa, \ourmethod{} improves accuracy by up to $2.56\%$. Finally, models under RNS-CKKS (AutoFHE) accelerate inference by $103\times$ and improve accuracy by $3.46\%$ compared to models under TFHE (\redsec{}). 

Our results demonstrate the effectiveness of automated search-based algorithms in navigating the large search space of adapting convolution neural networks for inference over FHE ciphertexts and discovering networks that Pareto-dominate manually designed ones. In summary, an integrated and automated design of polynomial approximations and homomorphic evaluation architecture is an effective and flexible approach for seamlessly adapting CNNs for inference on FHE ciphertexts.

\bibliographystyle{plain}
\bibliography{ref.bib}

\begin{thebibliography}{10}

\bibitem{beaver1995precomputing}
Donald Beaver.
\newblock Precomputing oblivious transfer.
\newblock In {\em Annual International Cryptology Conference}, pages 97--109, 1995.
\newblock \url{https://link.springer.com/chapter/10.1007/3-540-44750-4_8}.

\bibitem{bellare2012foundations}
Mihir Bellare, Viet~Tung Hoang, and Phillip Rogaway.
\newblock Foundations of garbled circuits.
\newblock In {\em Proceedings of the 2012 ACM Conference on Computer and Communications Security}, pages 784--796, 2012.
\newblock \url{https://eprint.iacr.org/2012/265.pdf}.

\bibitem{bengio2013estimating}
Yoshua Bengio, Nicholas L{\'e}onard, and Aaron Courville.
\newblock Estimating or propagating gradients through stochastic neurons for conditional computation.
\newblock {\em arXiv preprint arXiv:1308.3432}, 2013.
\newblock \url{https://arxiv.org/pdf/1308.3432.pdf}.

\bibitem{bossuat2021efficient}
Jean-Philippe Bossuat, Christian Mouchet, Juan Troncoso-Pastoriza, and Jean-Pierre Hubaux.
\newblock Efficient bootstrapping for approximate homomorphic encryption with non-sparse keys.
\newblock In {\em Annual International Conference on the Theory and Applications of Cryptographic Techniques}, pages 587--617, 2021.
\newblock \url{https://link.springer.com/chapter/10.1007/978-3-030-77870-5_21}.

\bibitem{brakerski2012fully}
Zvika Brakerski.
\newblock Fully homomorphic encryption without modulus switching from classical gapsvp.
\newblock In {\em Annual Cryptology Conference}, pages 868--886, 2012.
\newblock \url{https://link.springer.com/chapter/10.1007/978-3-642-32009-5_50}.

\bibitem{brutzkus2019low}
Alon Brutzkus, Ran Gilad-Bachrach, and Oren Elisha.
\newblock Low latency privacy preserving inference.
\newblock In {\em International Conference on Machine Learning}, pages 812--821, 2019.
\newblock \url{http://proceedings.mlr.press/v97/brutzkus19a/brutzkus19a.pdf}.

\bibitem{cheon2018bootstrapping}
Jung~Hee Cheon, Kyoohyung Han, Andrey Kim, Miran Kim, and Yongsoo Song.
\newblock Bootstrapping for approximate homomorphic encryption.
\newblock In {\em Annual International Conference on the Theory and Applications of Cryptographic Techniques}, pages 360--384, 2018.
\newblock \url{https://link.springer.com/chapter/10.1007/978-3-319-78381-9_14}.

\bibitem{cheon2018full}
Jung~Hee Cheon, Kyoohyung Han, Andrey Kim, Miran Kim, and Yongsoo Song.
\newblock A full {RNS} variant of approximate homomorphic encryption.
\newblock In {\em International Conference on Selected Areas in Cryptography}, pages 347--368, 2018.
\newblock \url{https://link.springer.com/chapter/10.1007/978-3-030-10970-7_16}.

\bibitem{cheon2019hybrid}
Jung~Hee Cheon, Minki Hhan, Seungwan Hong, and Yongha Son.
\newblock A hybrid of dual and meet-in-the-middle attack on sparse and ternary secret {LWE}.
\newblock {\em IEEE Access}, 7:89497--89506, 2019.
\newblock \url{https://ieeexplore.ieee.org/stamp/stamp.jsp?tp=&arnumber=8747481}.

\bibitem{cheon2017homomorphic}
Jung~Hee Cheon, Andrey Kim, Miran Kim, and Yongsoo Song.
\newblock Homomorphic encryption for arithmetic of approximate numbers.
\newblock In {\em International Conference on the Theory and Application of Cryptology and Information Security}, pages 409--437, 2017.
\newblock \url{https://link.springer.com/chapter/10.1007/978-3-319-70694-8_15}.

\bibitem{chillotti2020tfhe}
Ilaria Chillotti, Nicolas Gama, Mariya Georgieva, and Malika Izabach{\`e}ne.
\newblock {TFHE:} {F}ast fully homomorphic encryption over the torus.
\newblock {\em Journal of Cryptology}, 33(1):34--91, 2020.
\newblock \url{https://eprint.iacr.org/2018/421.pdf}.

\bibitem{chou2018faster}
Edward Chou, Josh Beal, Daniel Levy, Serena Yeung, Albert Haque, and Li~Fei-Fei.
\newblock {Faster CryptoNets:} {L}everaging sparsity for real-world encrypted inference.
\newblock {\em arXiv preprint arXiv:1811.09953}, 2018.
\newblock \url{https://arxiv.org/pdf/1811.09953.pdf}.

\bibitem{deb2002fast}
Kalyanmoy Deb, Amrit Pratap, Sameer Agarwal, and Tamt Meyarivan.
\newblock A fast and elitist multiobjective genetic algorithm: {NSGA-II}.
\newblock {\em IEEE Transactions on Evolutionary Computation}, 6(2):182--197, 2002.
\newblock \url{https://ieeexplore.ieee.org/stamp/stamp.jsp?tp=&arnumber=996017}.

\bibitem{devlin2018bert}
Jacob Devlin, Ming-Wei Chang, Kenton Lee, and Kristina Toutanova.
\newblock {BERT:} pre-training of deep bidirectional transformers for language understanding.
\newblock {\em arXiv preprint arXiv:1810.04805}, 2018.
\newblock \url{https://arxiv.org/abs/1810.04805}.

\bibitem{dosovitskiy2020image}
Alexey Dosovitskiy, Lucas Beyer, Alexander Kolesnikov, Dirk Weissenborn, Xiaohua Zhai, Thomas Unterthiner, Mostafa Dehghani, Matthias Minderer, Georg Heigold, Sylvain Gelly, et~al.
\newblock An image is worth 16x16 words: Transformers for image recognition at scale.
\newblock In {\em International Conference on Learning Representations}, 2020.
\newblock \url{https://openreview.net/pdf?id=YicbFdNTTy}.

\bibitem{fan2012somewhat}
Junfeng Fan and Frederik Vercauteren.
\newblock Somewhat practical fully homomorphic encryption.
\newblock {\em Cryptology ePrint Archive}, 2012.
\newblock \url{https://ia.cr/2012/144}.

\bibitem{fawzi2022discovering}
Alhussein Fawzi, Matej Balog, Aja Huang, Thomas Hubert, Bernardino Romera-Paredes, Mohammadamin Barekatain, Alexander Novikov, Francisco~J R~Ruiz, Julian Schrittwieser, Grzegorz Swirszcz, et~al.
\newblock Discovering faster matrix multiplication algorithms with reinforcement learning.
\newblock {\em Nature}, 610(7930):47--53, 2022.
\newblock \url{https://www.nature.com/articles/s41586-022%20-05172-4}.

\bibitem{folkerts2023redsec}
Lars~Wolfgang Folkerts, Charles Gouert, and Nektarios~Georgios Tsoutsos.
\newblock {REDsec}: Running encrypted discretized neural networks in seconds.
\newblock In {\em Network and Distributed System Security Symposium}, 2023.
\newblock \url{https://www.ndss-symposium.org/wp-content/uploads/2023/02/ndss2023_f34_paper.pdf}.

\bibitem{ghodsi2021circa}
Zahra Ghodsi, Nandan~Kumar Jha, Brandon Reagen, and Siddharth Garg.
\newblock Circa: Stochastic relus for private deep learning.
\newblock In {\em Advances in Neural Information Processing Systems}, volume~34, pages 2241--2252, 2021.
\newblock \url{https://proceedings.neurips.cc/paper/2021/file/11eba2991cc62daa4a85be5c0cfdae97-Paper.pdf}.

\bibitem{gilad2016cryptonets}
Ran Gilad-Bachrach, Nathan Dowlin, Kim Laine, Kristin Lauter, Michael Naehrig, and John Wernsing.
\newblock {CryptoNets}: Applying neural networks to encrypted data with high throughput and accuracy.
\newblock In {\em International Conference on Machine Learning}, pages 201--210, 2016.
\newblock \url{http://proceedings.mlr.press/v48/gilad-bachrach16.pdf}.

\bibitem{goldberg1991comparative}
David~E Goldberg and Kalyanmoy Deb.
\newblock A comparative analysis of selection schemes used in genetic algorithms.
\newblock In {\em Foundations of Genetic Algorithms}, volume~1, pages 69--93. Elsevier, 1991.
\newblock \url{https://www.sciencedirect.com/science/article/abs/pii/B9780080506845500082}.

\bibitem{goyal2017accurate}
Priya Goyal, Piotr Doll{\'a}r, Ross Girshick, Pieter Noordhuis, Lukasz Wesolowski, Aapo Kyrola, Andrew Tulloch, Yangqing Jia, and Kaiming He.
\newblock Accurate, large minibatch sgd: Training imagenet in 1 hour.
\newblock {\em arXiv preprint arXiv:1706.02677}, 2017.
\newblock \url{https://arxiv.org/abs/1706.02677}.

\bibitem{hao2022iron}
Meng Hao, Hongwei Li, Hanxiao Chen, Pengzhi Xing, Guowen Xu, and Tianwei Zhang.
\newblock Iron: Private inference on transformers.
\newblock In {\em Advances in Neural Information Processing Systems}, volume~35, pages 15718--15731, 2022.
\newblock \url{https://openreview.net/forum?id=deyqjpcTfsG}.

\bibitem{he2016deep}
Kaiming He, Xiangyu Zhang, Shaoqing Ren, and Jian Sun.
\newblock Deep residual learning for image recognition.
\newblock In {\em Proceedings of the IEEE Conference on Computer Vision and Pattern Recognition}, pages 770--778, 2016.
\newblock \url{https://openaccess.thecvf.com/content_cvpr_2016/papers/He_Deep_Residual_Learning_CVPR_2016_paper.pdf}.

\bibitem{hinton2015distilling}
Geoffrey Hinton, Oriol Vinyals, and Jeff Dean.
\newblock Distilling the knowledge in a neural network.
\newblock {\em arXiv preprint arXiv:1503.02531}, 2015.
\newblock \url{https://arxiv.org/abs/1503.02531}.

\bibitem{jacob2018quantization}
Benoit Jacob, Skirmantas Kligys, Bo~Chen, Menglong Zhu, Matthew Tang, Andrew Howard, Hartwig Adam, and Dmitry Kalenichenko.
\newblock Quantization and training of neural networks for efficient integer-arithmetic-only inference.
\newblock In {\em IEEE Conference on Computer Vision and Pattern Recognition}, 2018.
\newblock \url{https://openaccess.thecvf.com/content_cvpr_2018/papers/Jacob_Quantization_and_Training_CVPR_2018_paper.pdf}.

\bibitem{jha2021deepreduce}
Nandan~Kumar Jha, Zahra Ghodsi, Siddharth Garg, and Brandon Reagen.
\newblock {DeepReDuce}: {ReLU} reduction for fast private inference.
\newblock In {\em International Conference on Machine Learning}, pages 4839--4849, 2021.
\newblock \url{http://proceedings.mlr.press/v139/jha21a.html}.

\bibitem{jumper2021highly}
John Jumper, Richard Evans, Alexander Pritzel, Tim Green, Michael Figurnov, Olaf Ronneberger, Kathryn Tunyasuvunakool, Russ Bates, Augustin {\v{Z}}{\'\i}dek, Anna Potapenko, et~al.
\newblock Highly accurate protein structure prediction with alphafold.
\newblock {\em Nature}, 596(7873):583--589, 2021.
\newblock \url{https://www.nature.com/articles/s41586-021-03819-2}.

\bibitem{juvekar2018gazelle}
Chiraag Juvekar, Vinod Vaikuntanathan, and Anantha Chandrakasan.
\newblock {GAZELLE}: A low latency framework for secure neural network inference.
\newblock In {\em USENIX Security Symposium}, pages 1651--1669, 2018.
\newblock \url{https://www.usenix.org/system/files/conference/usenixsecurity18/sec18-juvekar.pdf}.

\bibitem{knott2021crypten}
Brian Knott, Shobha Venkataraman, Awni Hannun, Shubho Sengupta, Mark Ibrahim, and Laurens van~der Maaten.
\newblock {CrypTen}: Secure multi-party computation meets machine learning.
\newblock In {\em Advances in Neural Information Processing Systems}, volume~34, pages 4961--4973, 2021.
\newblock \url{https://proceedings.neurips.cc/paper/2021/file/2754518221cfbc8d25c13a06a4cb8421-Paper.pdf}.

\bibitem{krizhevsky2009learning}
Alex Krizhevsky, Geoffrey Hinton, et~al.
\newblock Learning multiple layers of features from tiny images.
\newblock 2009.
\newblock \url{https://www.cs.toronto.edu/~kriz/learning-features-2009-TR.pdf}.

\bibitem{lee2021minimax}
Eunsang Lee, Joon-Woo Lee, Young-Sik Kim, and Jong-Seon No.
\newblock Minimax approximation of sign function by composite polynomial for homomorphic comparison.
\newblock {\em IEEE Transactions on Dependable and Secure Computing}, 2021.
\newblock \url{https://ieeexplore.ieee.org/stamp/stamp.jsp?tp=&arnumber=9517029}.

\bibitem{lee2022low}
Eunsang Lee, Joon-Woo Lee, Junghyun Lee, Young-Sik Kim, Yongjune Kim, Jong-Seon No, and Woosuk Choi.
\newblock Low-complexity deep convolutional neural networks on fully homomorphic encryption using multiplexed parallel convolutions.
\newblock In {\em International Conference on Machine Learning}, pages 12403--12422, 2022.
\newblock \url{https://proceedings.mlr.press/v162/lee22e/lee22e.pdf}.

\bibitem{lee2022privacy}
Joon-Woo Lee, HyungChul Kang, Yongwoo Lee, Woosuk Choi, Jieun Eom, Maxim Deryabin, Eunsang Lee, Junghyun Lee, Donghoon Yoo, Young-Sik Kim, et~al.
\newblock Privacy-preserving machine learning with fully homomorphic encryption for deep neural network.
\newblock {\em IEEE Access}, 10:30039--30054, 2022.
\newblock \url{https://ieeexplore.ieee.org/stamp/stamp.jsp?tp=&arnumber=9734024}.

\bibitem{lee2020optimal}
Joon-Woo Lee, Eunsang Lee, Yongwoo Lee, Young-Sik Kim, and Jong-Seon No.
\newblock High-precision bootstrapping of rns-ckks homomorphic encryption using optimal minimax polynomial approximation and inverse sine function.
\newblock In {\em Annual International Conference on the Theory and Applications of Cryptographic Techniques}, pages 618--647, 2021.
\newblock \url{https://eprint.iacr.org/2020/552.pdf}.

\bibitem{lee2021precise}
Junghyun Lee, Eunsang Lee, Joon-Woo Lee, Yongjune Kim, Young-Sik Kim, and Jong-Seon No.
\newblock Precise approximation of convolutional neural networks for homomorphically encrypted data.
\newblock {\em arXiv preprint arXiv:2105.10879}, 2021.
\newblock \url{https://arxiv.org/pdf/2105.10879.pdf}.

\bibitem{leshno1993multilayer}
Moshe Leshno, Vladimir~Ya Lin, Allan Pinkus, and Shimon Schocken.
\newblock Multilayer feedforward networks with a nonpolynomial activation function can approximate any function.
\newblock {\em Neural Networks}, 6(6):861--867, 1993.
\newblock \url{https://www.sciencedirect.com/science/article/pii/S0893608005801315}.

\bibitem{li2016ternary}
Fengfu Li, Bin Liu, Xiaoxing Wang, Bo~Zhang, and Junchi Yan.
\newblock Ternary weight networks.
\newblock {\em arXiv preprint arXiv:1605.04711}, 2016.
\newblock \url{https://arxiv.org/pdf/1605.04711.pdf}.

\bibitem{liu2018darts}
Hanxiao Liu, Karen Simonyan, and Yiming Yang.
\newblock Darts: Differentiable architecture search.
\newblock In {\em International Conference on Learning Representations}, 2018.
\newblock \url{https://openreview.net/pdf?id=S1eYHoC5FX}.

\bibitem{liu2017oblivious}
Jian Liu, Mika Juuti, Yao Lu, and Nadarajah Asokan.
\newblock Oblivious neural network predictions via minionn transformations.
\newblock In {\em Proceedings of the 2017 ACM SIGSAC Conference on Computer and Communications Security}, pages 619--631, 2017.
\newblock \url{https://eprint.iacr.org/2017/452.pdf}.

\bibitem{lou2021hemet}
Qian Lou and Lei Jiang.
\newblock {HEMET}: A homomorphic-encryption-friendly privacy-preserving mobile neural network architecture.
\newblock In {\em International Conference on Machine Learning}, pages 7102--7110, 2021.
\newblock \url{http://proceedings.mlr.press/v139/lou21a/lou21a.pdf}.

\bibitem{lou2020safenet}
Qian Lou, Yilin Shen, Hongxia Jin, and Lei Jiang.
\newblock {SAFENet}: A secure, accurate and fast neural network inference.
\newblock In {\em International Conference on Learning Representations}, 2020.
\newblock \url{https://openreview.net/pdf?id=Cz3dbFm5u-}.

\bibitem{ma2018survey}
Xiaoliang Ma, Xiaodong Li, Qingfu Zhang, Ke~Tang, Zhengping Liang, Weixin Xie, and Zexuan Zhu.
\newblock A survey on cooperative co-evolutionary algorithms.
\newblock {\em IEEE Transactions on Evolutionary Computation}, 23(3):421--441, 2018.
\newblock \url{https://ieeexplore.ieee.org/stamp/stamp.jsp?tp=&arnumber=8454482}.

\bibitem{mei2016competitive}
Yi~Mei, Mohammad~Nabi Omidvar, Xiaodong Li, and Xin Yao.
\newblock A competitive divide-and-conquer algorithm for unconstrained large-scale black-box optimization.
\newblock {\em ACM Transactions on Mathematical Software}, 42(2):1--24, 2016.
\newblock \url{https://dl.acm.org/doi/pdf/10.1145/2791291}.

\bibitem{mishra2020delphi}
Pratyush Mishra, Ryan Lehmkuhl, Akshayaram Srinivasan, Wenting Zheng, and Raluca~Ada Popa.
\newblock {Delphi}: A cryptographic inference service for neural networks.
\newblock In {\em USENIX Security Symposium}, pages 2505--2522, 2020.
\newblock \url{https://www.usenix.org/system/files/sec20-mishra_0.pdf}.

\bibitem{park2022aespa}
Jaiyoung Park, Michael~Jaemin Kim, Wonkyung Jung, and Jung~Ho Ahn.
\newblock {AESPA}: Accuracy preserving low-degree polynomial activation for fast private inference.
\newblock {\em arXiv preprint arXiv:2201.06699}, 2022.
\newblock \url{https://arxiv.org/abs/2201.06699}.

\bibitem{rathee2021sirnn}
Deevashwer Rathee, Mayank Rathee, Rahul Kranti~Kiran Goli, Divya Gupta, Rahul Sharma, Nishanth Chandran, and Aseem Rastogi.
\newblock {SiRnn:} a math library for secure rnn inference.
\newblock In {\em IEEE Symposium on Security and Privacy}, pages 1003--1020, 2021.
\newblock \url{https://eprint.iacr.org/2021/459.pdf}.

\bibitem{rauf2021adaptive}
Hafiz~Tayyab Rauf, Waqas Haider~Khan Bangyal, and M~Ikramullah Lali.
\newblock An adaptive hybrid differential evolution algorithm for continuous optimization and classification problems.
\newblock {\em Neural Computing and Applications}, 33(17):10841--10867, 2021.
\newblock \url{https://link.springer.com/article/10.1007/s00521-021-06216-y}.

\bibitem{schrittwieser2020mastering}
Julian Schrittwieser, Ioannis Antonoglou, Thomas Hubert, Karen Simonyan, Laurent Sifre, Simon Schmitt, Arthur Guez, Edward Lockhart, Demis Hassabis, Thore Graepel, et~al.
\newblock Mastering atari, go, chess and shogi by planning with a learned model.
\newblock {\em Nature}, 588(7839):604--609, 2020.
\newblock \url{https://www.nature.com/articles/s41586-020-03051-4}.

\bibitem{seal3.6}
SEAL.
\newblock Microsoft {SEAL} (3.6).
\newblock {\em Microsoft Research}, 2020.
\newblock \url{https://github.com/Microsoft/SEAL}.

\bibitem{silver2018general}
David Silver, Thomas Hubert, Julian Schrittwieser, Ioannis Antonoglou, Matthew Lai, Arthur Guez, Marc Lanctot, Laurent Sifre, Dharshan Kumaran, Thore Graepel, et~al.
\newblock A general reinforcement learning algorithm that masters chess, shogi, and go through self-play.
\newblock {\em Science}, 362(6419):1140--1144, 2018.
\newblock \url{https://www.science.org/doi/full/10.1126/science.aar6404}.

\bibitem{srinivas1994muiltiobjective}
Nidamarthi Srinivas and Kalyanmoy Deb.
\newblock Muiltiobjective optimization using nondominated sorting in genetic algorithms.
\newblock {\em Evolutionary Computation}, 2(3):221--248, 1994.
\newblock \url{https://ieeexplore.ieee.org/abstract/document/6791727}.

\bibitem{tan2021efficientnetv2}
Mingxing Tan and Quoc Le.
\newblock {EfficientNetV2}: Smaller models and faster training.
\newblock In {\em International Conference on Machine Learning}, volume 139, pages 10096--10106, 2021.

\bibitem{tfhe1.1}
TFHE.
\newblock Tfhe v1.1.
\newblock 2020.
\newblock \url{https://github.com/tfhe/tfhe}.

\bibitem{tolstikhin2021mlp}
Ilya~O Tolstikhin, Neil Houlsby, Alexander Kolesnikov, Lucas Beyer, Xiaohua Zhai, Thomas Unterthiner, Jessica Yung, Andreas Steiner, Daniel Keysers, Jakob Uszkoreit, et~al.
\newblock {MLP-Mixer}: An all-mlp architecture for vision.
\newblock In {\em Advances in Neural Information Processing Systems}, volume~34, pages 24261--24272, 2021.
\newblock \url{https://proceedings.neurips.cc/paper/2021/file/cba0a4ee5ccd02fda0fe3f9a3e7b89fe-Paper.pdf}.

\bibitem{vaswani2017attention}
Ashish Vaswani, Noam Shazeer, Niki Parmar, Jakob Uszkoreit, Llion Jones, Aidan~N Gomez, {\L}ukasz Kaiser, and Illia Polosukhin.
\newblock Attention is all you need.
\newblock In {\em Advances in Neural Information Processing Systems}, volume~30, 2017.
\newblock \url{https://proceedings.neurips.cc/paper/2017/file/3f5ee243547dee91fbd053c1c4a845aa-Paper.pdf}.

\bibitem{weng2021large}
Lilian Weng.
\newblock How to train really large models on many gpus?
\newblock {\em lilianweng.github.io}, 2021.
\newblock \url{https://lilianweng.github.io/posts/2021-09-25-train-large/}.

\bibitem{yang2008large}
Zhenyu Yang, Ke~Tang, and Xin Yao.
\newblock Large scale evolutionary optimization using cooperative coevolution.
\newblock {\em Information Sciences}, 178(15):2985--2999, 2008.
\newblock \url{https://www.sciencedirect.com/science/article/pii/S002002550800073X}.

\bibitem{yao1986generate}
Andrew Chi-Chih Yao.
\newblock How to generate and exchange secrets.
\newblock In {\em 27th Annual Symposium on Foundations of Computer Science}, pages 162--167, 1986.
\newblock \url{https://ieeexplore.ieee.org/stamp/stamp.jsp?tp=&arnumber=4568207}.

\bibitem{zoph2016neural}
Barret Zoph and Quoc~V Le.
\newblock Neural architecture search with reinforcement learning.
\newblock {\em arXiv preprint arXiv:1611.01578}, 2016.
\newblock \url{https://arxiv.org/abs/1611.01578}.

\end{thebibliography}

\appendix
\section{\fhemp~under RNS-CKKS \label{appendix:mpcnn}}

\fhemp~\cite{lee2022low} is the state-of-the-art framework for homomorphically evaluating deep CNNs on encrypted data under RNS-CKKS with high accuracy. Its salient features are as follows.

\textbf{(1)} \textsc{Compact Packing:} All channels of a tensor are packed into a single ciphertext via multiplexed packing. Furthermore, multiplexed parallel (MP) convolution was proposed to process the ciphertext efficiently. Multiplexed packing can effectively avoid wasting slots due to strided convolutions. 

\textbf{(2)} \textsc{Homomorphic Evaluation Architecture:} To refresh zero-level ciphertexts, bootstrapping operations are placed after every ConvBN (as shown in Figure~\ref{fig:connect}), except for the first one. This hand-crafted homomorphic evaluation architecture for ResNets is determined by the choice of cryptographic parameters, the level consumption of operations, and ResNet's architectures.

\textbf{(3)} \textsc{AppReLU:} It replaces all ReLUs with the same high-degree Minimax composite polynomial~\cite{lee2021minimax,lee2021precise} of degrees $\{15,15,27\}$. By noting that $\mathrm{ReLU}(x)=x\cdot(0.5+0.5\cdot \mathrm{sgn}(x))$, where $\mathrm{sgn}(x)$ is the sign function, the approximated ReLU (AppReLU) is modeled as $\mathrm{AppReLU}(x)=x\cdot(0.5+0.5\cdot p_{\alpha}(x)),x\in[-1,1]$. $p_{\alpha}(x)$ is the composite Minimax polynomial.  The precision $\alpha$ is defined as $|p_{\alpha}(x)-\mathrm{sgn(x)}|\leq 2^{-\alpha}$. AppReLU is expanded to arbitrary domains $x\in[-B, B]$ of pre-activations in CNNs by scaling it as $B\cdot\mathrm{AppReLU}(x / B)$. $B$ can be estimated on the training dataset.

\textbf{(4)} \textsc{Cryptographic Parameters:} \fhemp~sets $N=2^{16}$, $L=30$ and Hamming weight $h=192$. Base modulus, special modulus, and bootstrapping modulus are set to 51 bits, while default modulus is set to 46 bits. These cryptographic parameters satisfy 128 bits of security~\cite{cheon2019hybrid}. 

\textbf{(5)} \textsc{Multiplicative Depth: } 
The multiplicative depth of Bootstrapping (i.e., $K$), AppReLU, ConvBN, DownSampling, AvgPool, FC layers are 14, 14, 2, 1, 1, 1 respectively. Statistically, when using \fhemp~to homomorphically evaluate ResNet-18/32/44 on CIFAR10 or CIFAR100, AppReLUs consume $\sim50\%$ of total levels, and bootstrapping operations consume $>70\%$ of latency.

\section{\aespa~under RNS-CKKS \label{appendix:aespa}}
\aespa~\cite{park2022aespa} proposes HerPN to replace ReLU and Batchnorm. HerPN is the approximation of ReLU using Hermite polynomials followed by basis-wise normalization. The Hermite approximation is  $f(x)=\sum_{i=0}^\infty \hat{f_i}h_i(x)$ where normalized probabilist's Hermite polynomial $h_i(x)=H_n(x)/\sqrt{n!}$ and coefficient $\hat{f_i}=\int_{-\infty}^\infty f(x) \overline{h_i(x)} e^{-x^2/2}$. Given tensor $\bm{x}\in\mathbb{R}^{N\times C\times H\times W}$, HerPN is defined as:
\begin{equation}
    f(\bm{x})=\bm{\gamma} \sum_{i=0}^d \hat{f_i}\frac{h_i(\bm{x})-\bm{\mu}_i}{\sqrt{\bm{\sigma}_i^2+\epsilon}} + \bm{\beta}
\end{equation}
where mean $\bm{\mu}_i\in\mathbb{R}^C$ and standard variance $\bm{\sigma}_i\in\mathbb{R}^C$ are computed over output of each Hermite polynomial, $h_i(\bm{x})$. $\bm{\gamma}\in\mathbb{R^C}$ and $\bm{\beta}\in\mathbb{R}^C$ are learnable parameters. \aespa~uses the first three Hermite bases, \ie $d=2$. Hermite bases and coefficients are: 
\begin{align}
h_0(x)&=1 & h_1(x)&=x  &  h_2(x)&=\frac{x^2-1}{\sqrt{2}}\\
\hat{f_0}&=\frac{1}{\sqrt{2\pi}} & \hat{f_1}&=\frac{1}{2} & \hat{f_3}&=\frac{1}{\sqrt{4\pi}}
\end{align}
HerPN can be cast as a second-degree polynomial to reduce multiplicative depth, namely:
\begin{multline}
f(\bm{x}) = \frac{\bm{\gamma}}{\sqrt{8\pi(\bm{\sigma}_2^2+\epsilon)}}\bm{x}^2 + \frac{\bm{\gamma}}{2\sqrt{\bm{\sigma}_1^2+\epsilon}}\bm{x}+\bm{\beta}+\\
\bm{\gamma}\left[ \frac{1-\bm{\mu}_0}{\sqrt{2\pi(\bm{\sigma}_0^2+\epsilon)}}
-\frac{\bm{\mu}_1}{2\sqrt{\bm{\sigma}_1^2+\epsilon}}-\frac{1+\sqrt{2}\bm{\mu}_2}{\sqrt{8\pi(\bm{\sigma}_2^2+\epsilon)}} \right]
\label{eq:herpn_poly}
\end{multline}
From \Eqref{eq:herpn_poly}, the depth of HerPN is 2. Hence, 4 Conv-HerPN layers should be followed by one bootstrapping operation. 
\aespa~was originally proposed for secure MPC~\cite{park2022aespa}. Our paper adopts \aespa{} as a low-degree baseline under RNS-CKKS. We design the homomorphic evaluation architecture of \aespa~under RNS-CKKS as shown in Figure~\ref{fig:connect}. We build its C++ implementation on top of \fhemp{} by using its implementations of Conv, BN, Downsample, AvgPool, and FC layers.

\end{document}